\documentclass[runningheads]{llncs}
\usepackage{graphicx}
\usepackage{amsmath,amssymb} 
\usepackage[usenames, dvipsnames]{color}
\usepackage{wrapfig}
\usepackage{multirow}
\usepackage{subfig}
\usepackage{url}
\usepackage[normalem]{ulem}

\newif\ifdrafting
\draftingtrue 
\draftingfalse 

\begin{document}
\pagestyle{headings}
\mainmatter

\definecolor{darkgreen}{RGB}{0,127,0}
\definecolor{darkblue}{RGB}{0,0,127}
\definecolor{darkmagenta}{RGB}{127,0,127}
\definecolor{darkred}{RGB}{127,0,0}
\definecolor{darkcyan}{RGB}{0,127,127}

\ifdrafting
  \newcommand{\QG} [1] {\textcolor{darkgreen}{[QG: #1]}} 
  \newcommand{\OG} [1] {\textcolor{darkblue}{[OG: #1]}} 
  \newcommand{\IF} [1] {\textcolor{darkmagenta}{[IF: #1]}} 
  \newcommand{\TZ} [1] {\textcolor{darkred}{[TZ: #1]}} 
  \newcommand{\checkthis}[1]{\textcolor{darkcyan}{[check this $\rightarrow$] #1}} 
  \newcommand{\new} [1] {\textcolor{darkgreen}{#1}} 
  \newcommand{\old} [1] {\sout{#1}}
\else
  \newcommand{\OG} [1] {}
  \newcommand{\PW} [1] {}
  \newcommand{\NW} [1] {#1} 
  \newcommand{\JG} [1] {}
  \newcommand{\JK} [1] {}
  \newcommand{\TODO} [1] {}
  \newcommand{\checkthis}[1]{#1}
  \newcommand{\old} [1] {}
  \newcommand{\new} [1] {#1}
\fi

\title{Tackling 3D ToF Artifacts Through Learning and the FLAT Dataset} 

\titlerunning{Tackling 3D ToF Artifacts}

\authorrunning{Q. Guo, I. Frosio, O. Gallo, T. Zickler, and J. Kautz}

\author{Qi Guo\textsuperscript{a,b}, Iuri Frosio\textsuperscript{a}, Orazio Gallo\textsuperscript{a}, Todd Zickler\textsuperscript{b}, Jan Kautz\textsuperscript{a}}


\institute{
\begin{tabular}{cc}
\textsuperscript{a}NVIDIA, &\textsuperscript{b}Harvard SEAS 
\end{tabular}
}

\maketitle

\begin{abstract}
Scene motion, multiple reflections, and sensor noise introduce artifacts in the depth reconstruction performed by time-of-flight cameras.
We propose a two-stage, deep-learning approach to address all of these sources of artifacts simultaneously.
We also introduce FLAT, a synthetic dataset of 2000 ToF measurements that capture all of these nonidealities, \new{and allows to simulate different camera hardware.} 
\old{and can be used to simulate different hardware.}
Using the Kinect 2 camera as a baseline, we show improved reconstruction errors \new{over state-of-the-art methods,} on \new{both} simulated and real data\old{, as compared with state-of-the-art methods}.
\end{abstract}

\section{Introduction}\label{sec:intro}

Depth estimation is central to several computer vision applications. Among the many existing strategies for extracting a scene's 3D information, Time-of-Flight (ToF) cameras are particularly popular due to their robustness and affordability.

ToF cameras leverage the relation between an object's distance from the sensor and the amount of time required for photons to travel to that object and back. In particular, \old{the widely used}Amplitude-Modulated Continuous-Wave (AMCW) cameras emit a periodic light signal and measure its phase delay upon return: the phase delay offers an estimate of \old{time-of-flight}\new{the time of flight} and, in turn, of depth. Any implementation of this approach requires several careful considerations.


\begin{figure}[t]
\begin{center}
\includegraphics[width=\textwidth]{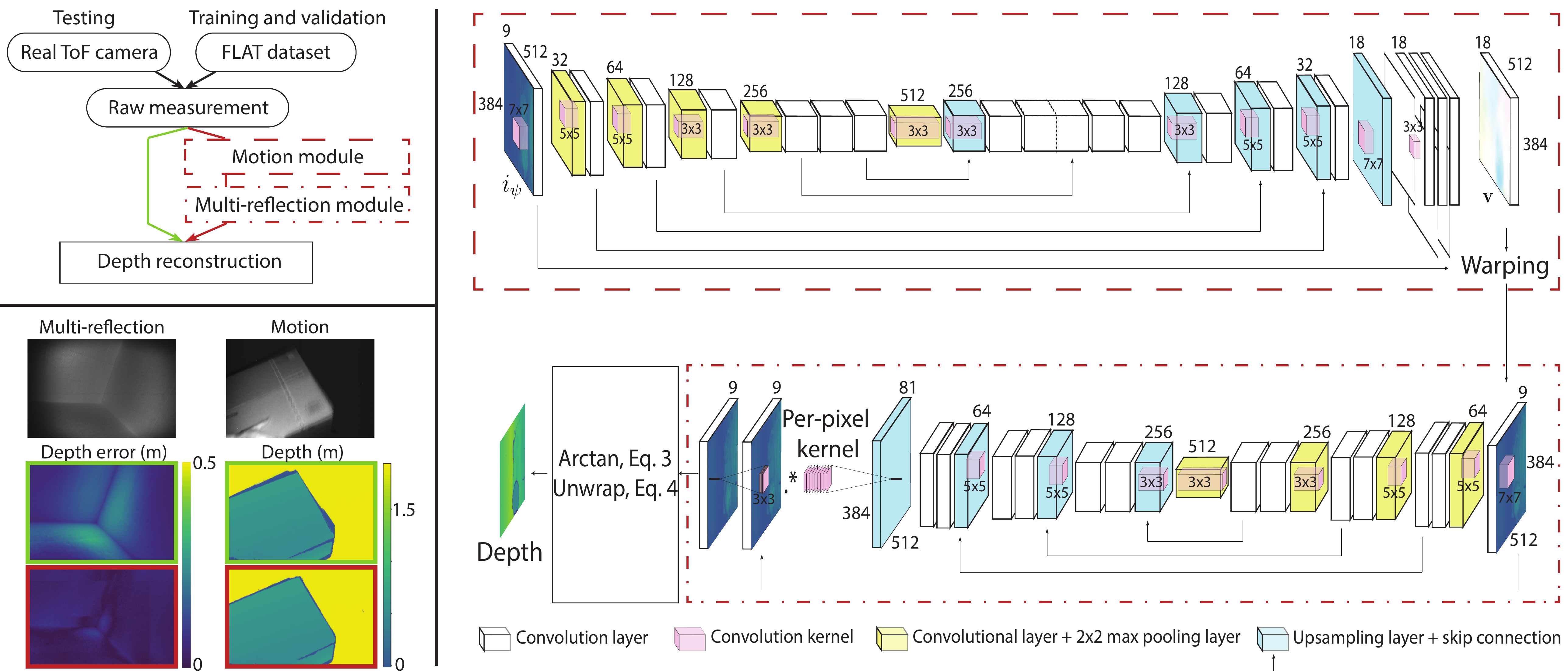}
\caption{The traditional ToF processing pipeline ({\color{ForestGreen}green}, libfreenect2~\cite{libfreenect2} as an example) and the proposed framework ({\color{Red}red}).
The lower left panel shows artifacts generated by MPI and motion, that show up respectively as deformation close to corners and spikes or missing data close to the boundaries of moving objects in the traditional flowchart ({\color{ForestGreen}green}).
These artifacts are greatly reduced by the proposed framework ({\color{Red}red}), which is based on the introduction of the motion and multi-reflection modules depicted in the right part.}
\label{fig:teaser}
\end{center}
\end{figure}

First and foremost, the phase delay wraps at distances that correspond to multiples of the modulation period.
A common approach is to combine information from different modulation frequencies: longer modulation periods extend the unambiguous range, while shorter periods allow resolving finer details.
Using multiple frequencies, however, is not without consequences.
The dynamic parts of the scene may be displaced between sequential measurements at different frequencies, causing depth estimation errors\footnote{Some cameras use spatial multiplexing instead, and synchronize neighboring pixels with specific emission frequencies, thus sacrificing spatial resolution.}
that are particularly strong along the depth discontinuities. These artifacts can be identified and removed, at the cost of missing depth information at the corresponding pixels.

Another consideration is multi-path light transport. In addition to the direct emitter-object-sensor path, light may follow other paths and bounce multiple times before being recorded by one pixel. This phenomenon, called multiple-path interference (MPI), causes biased estimates of depth.
Several methods attempt to attenuate the effects of MPI~\cite{Bha16,Fei16,marco2017deeptof,Mut15,Son16}, some employing multiple modulation frequencies~\cite{Bha16,Fei16}\new{, which}\old{that}, as discussed above, can introduce motion artifacts.
These could be reduced in theory by increasing the capture speed, but shorter exposure times may lead to a lower signal-to-noise ratio because of \old{the}shot noise.


The delicate trade-offs between phase-unwrapping, motion, sensor noise, and MPI, have been studied in a variety of ways, with several methods attempting to attenuate the effects of \old{MPI~\cite{Bha16,Fei16,marco2017deeptof,Mut15,Son16}, motion artifacts~\cite{Lee14,Got14}, or shot noise~\cite{Len13}.}
\new{the different sources of artifacts \emph{independently}.}
\OG{We can revert this back, but I feel that it's an unnecessary repetition--we just talked about these things above.}
\old{However, and possibly because of the lack of extensive datasets, there have been no attempts to tackle them jointly by learning, to the best of our knowledge.}
\new{To the best of our knowledge, however, there have been no attempts to tackle them jointly by learning, possibly because of the lack of large datasets.}

We introduce a learning-based approach to tackle dynamic scenes, MPI, and shot noise \emph{simultaneously}. Our two-stage architecture (Fig.~\ref{fig:teaser}) operates directly on the raw measurements of a multi-frequency ToF sensor, and produces improved measurements that are compatible with  standard equations for phase-unwrapping and conversion to depth. \old{Our}\new{The} first stage is an encoder-decoder architecture~\cite{Dos15} that attenuates motion artifacts by spatially warping the raw measurements onto a common 2D reference frame. This increases the number of pixels whose depth can be reliably estimated, especially near the boundaries of \old{the}moving objects. The second stage is a kernel predicting network~\cite{Bako17} that attenuates MPI and shot noise. Because ground truth depth for real-world scenes is difficult to capture, we introduce FLAT (Flexible, Large, Augmentable, ToF dataset), a synthetic dataset for training and evaluation, which allows \old{accurately simulating}\new{to accurately simulate} the raw measurements of different AMCW ToF cameras (including the Kinect 2) in the presence of MPI artifacts and shot noise; FLAT data can be augmented to approximate motion artifacts as well.    
Our contributions are:

{\bf{(i)}} FLAT, a large, synthetic dataset of ToF measurements that simulates MPI, motion artifacts, shot noise, and different camera response functions.

{\bf{(ii)}} A Deep Neural Network (DNN) architecture for attenuating motion, MPI, and shot noise artifacts that can be trained both in the raw measurement or depth domain.

{\bf{(iii)}} A thorough validation, including an ablation study and a comparison with state-of-the-art algorithms for reducing MPI and motion artifacts.  

{\bf{(iv)}} A complete characterization of the Kinect 2 camera, including its camera response function and sensor noise characteristics.

Our DNN model, dataset and characterization of the Kinect 2 are available at  \url{http://research.nvidia.com/publication/2018-09_Tackling-3D-ToF}.

\section{Related Work}\label{sec:related}

Several works \old{in literature}separately reduce artifacts due to MPI, motion, or shot noise in AMCW ToF imaging.
We group them into four categories.
%
%

{\bf{Measurement noise reduction.}}
Raw ToF measurements suffer from both systematic and random noise~\cite{Jun14}. Systematic errors are often associated with imperfect sinusoidal modulations and can be reduced through calibration~\cite{Jun14,Jar17,Fer15}. Shot noise and other types of random noise are typically addressed through bilateral filtering of the raw measurements, the depth map, or both sometimes using other images for guidance~\cite{Len13}. The performance of these approaches is generally satisfactory, and any system that intends to replace them, ours included, should not perform noticeably worse. 

{\bf{Motion artifacts reduction.}} Motion artifacts occur when objects move and ToF raw measurements are captured sequentially. Gottfried et al.~\cite{Got14} identify three ways to attenuate them: reduce the number of sequential measurements; detect and correct the regions affected by motion, both in the raw measurement domain and in the depth domain; or estimate the 2D motion fields between raw 2D measurement maps and apply corrective spatial warping.
One way to detect affected pixels in raw measurements is by checking the \emph{Plus} and \emph{Minus rules}~\cite{Lee14,Han12} that derive from physical constraints on the light measured in a static scene. After detection, pixels affected by motion blur can be corrected, for example, by interpolation~\cite{Lee14}. Object motion also affects the frequency of reflected signals due to Doppler effect; however, to use the frequency shift to measure object motion requires a higher number of measurements and extensive processing~\cite{Heide2015}.

{\bf{Physics-based MPI reduction.}}
Algorithms for recovering depth from ToF correlations usually\old{ begin by assuming} \new{assume} the pure measurement of direct, single-bounce (emitter-surface-sensor) light paths. In practice, many photons bounce multiple times, causing ``erroneous'' measurements~\cite{Bha16,Jar17}. 
\old{If multiple modulation frequencies are used, the problem can be tackled by separately processing each pixel of the raw measurements in the temporal Fourier domain.}
\new{If multiple modulation frequencies are used, the problem can be tackled by processing the temporal change of each pixel of the raw measurements in the Fourier domain.}
For instance, in the absence of noise, $K$ interfering paths can be resolved by $2K+1$ frequency measurements~\cite{Fei16}. 
Other techniques for the per-pixel temporal processing of raw ToF measurements include Prony's method, the matrix pencil method, orthogonal matching pursuit, EPIRIT/MUSIC, and atomic norm regularization~\cite{Bha16}. 
Freedman et al. propose a real-time temporal processing technique that uses per-pixel optimization based on a light transport model with sparse and low-rank components~\cite{Fre14}.
Phasor Imaging exploits the fact that the effects of MPI are diminished at much higher modulation frequencies, and shows that simple temporal processing can succeed with as few as two such frequencies, albeit with a reduced unambiguous working range~\cite{Gup15}. 


{\bf{Learning-based MPI reduction.}}
The difficulty of modeling MPI analytically make\new{s} machine learning an enticing alternative for its reduction. However, one obstacle is the lack of large, physically-accurate datasets for training, which are difficult to capture~\cite{Nai12} and prohibitively expensive to simulate, until recently.
Marco et al. use an encoder to learn a mapping from captured ToF measurements to a representation of (MPI-corrupted) depth, and then combine this with a limited number of simulated, direct-only ToF measurements to train a decoder that produces MPI-corrected depth maps~\cite{marco2017deeptof}. 
Mutny et al. focus on corners of different materials and use a dataset of such corners to train a random forest with hand-crafted features~\cite{Mut15}.
A different strategy is taken by Son et al., who use a robotic arm and structured light system to capture ToF measurements with registered ground-truth depth~\cite{Son16}. They then train two neural networks to correct depth and refine edges through geodesic filtering.

We leverage the availability of computational power and advances in transient rendering~\cite{Jar14,Jar17} to synthesize a training dataset\old{,} 
sufficiently large and diverse to explore a much larger class of learned models. In addition to physically-accurate MPI effects, the dataset provides realistic shot noise, supports augmentation with approximate motion models, and allows \new{for} efficient generation of raw measurements from AMCW ToF sensors with arbitrary modulation signals.

\section{Time-of-Flight Camera Models}\label{sec:model}
In this section we first review the \old{math behind}\new{theory of} ToF reconstruction in the ideal case\old{with no MPI, motion, or shot noise artifacts}. 
We show that equations for depth reconstruction are differentiable, which allows \new{for backpropagation.}
\old{ backpropagation of the error through them.}
We then show the effect of MPI and motion on depth reconstruction, which help\new{s} framing the learning problem and defining the important factors 
for training.
We leverage these elements to train a neural network, working in the domain of the raw measurements and before unwrapping, aimed at correcting these artifacts.
The section is closed by an accurate characterization of the Kinect 2, which is our hardware testbed; this serves to produce accurate simulations for training and reduce the shift between synthetic and real data.
All the math details for the section are in the Supplementary.

\subsection{Ideal Camera Model}

An AMCW ToF camera illuminates the scene with an amplitude-modulated, sinusoidal signal $g(t)=g_1\text{cos}(\omega t)+g_0$, \cite{Heide2015}. If camera and light source are co-located, and the scene static, the signal coming back to a pixel is 
\begin{equation}
    s\left(t\right)= \int_{-\infty}^{t} a(\tau)\text{cos}(\omega t-2\omega\tau) d\tau,
\end{equation}
where $a(\tau)$ is the scene response function, i.e. the signal reaching the pixel at time $\tau$.
In an ideal scenario, the light is reflected once and the scene response is an impulse, $a(\tau) = a \delta(\tau - \tau_0)$.
The travel time $\tau_0$ directly translates to depth.

A \emph{homodyne} ToF camera modulates the incident signal with a phase-delayed reference signal at the same frequency, $b\cos\left(\omega t-\psi\right)$. The exposure time of the camera is usually set to $T >> 2\pi/\omega$. Simple trigonometry allows writing the raw correlation measurement $i_{\psi,\omega}$ as:
\begin{equation}
    i_{\psi,\omega} = \int_{-T/2}^{T/2} s(t) b\cos\left(\omega t-\psi\right) dt \approx a(\tau_0)b\cos(\psi-2\omega \tau_0) =
af_{\psi,\omega}(\tau_0),
    \label{eqn:itp}
\end{equation}
where we denote $f_{\psi,\omega}(\tau)=b\cos(\psi-2\omega \tau)$ as the {\it camera function}. Using raw measurements captured at a single frequency $\omega$ and $K\ge2$ phases $\vec{\psi}=(\psi_1,\ldots,\psi_K)$, the depth $d$ can be recovered at each pixel as:
\begin{equation}
     d = c / (2\omega) ~
      \arctan \left[(\sin\vec{\psi}\cdot \vec{i}_{\vec{\psi},\omega}) / (\cos\vec{\psi}\cdot \vec{i}_{\vec{\psi},\omega})\right],
     \label{eq:recon}
\end{equation}
where $\vec{i}_{\vec{\psi},\omega}$ is the $K$-vector of per-phase measurements.

However, $d$ wraps for depths larger than $ \pi c/\omega$, and additional measurements at $L\ge2$ different frequencies $\vec{\omega}=(\omega_1,\ldots,\omega_L)$ are required. Denoting the measurements at frequency $\omega_l$ as $\vec{i}_{\vec{\psi},\omega_l}$, an analytical expression for $d$ was provided by Goshov and Solodkin~\cite{gushov1991automatic} based on the Chinese remainder theorem:
\begin{equation}
d = \sum_{\vec{\omega}} A_l(\vec{\omega})  \arctan \left[(\sin\vec{\psi}\cdot \vec{i}_{\vec{\psi},\omega_l}) / (\cos\vec{\psi}\cdot \vec{i}_{\vec{\psi},\omega_l})\right]
 +  B(\vec{\omega}),
 \label{eq:gs}
\end{equation}
where $\{A_l(\vec{\omega})\}_{l=1,\ldots,L}$ and $B(\vec{\omega})$ are constants. Based on Eq.~\ref{eq:gs}, one can easily obtain the derivative $\partial{d}/\partial{\vec{i}_{\vec{\psi},\omega_l}}$, \new{which makes it possible to backpropagate the error on $d$ and to perform end-to-end training.}\old{which makes backpropagating the error on $d$ (and therefore end-to-end training of our neural network) possible.}

\subsection{The Impact of Multiple Paths}
\label{sec:mr}
In a realistic scenario, the signal that reaches the sensor is corrupted by multiple light paths that undergo different reflection events and have different path lengths.
This means that the scene response $a(\tau)$ is not an impulse anymore, as it measures the arrival of the light reaching a pixel from all the possible light paths that connect it to the emitter.
In this case, Eq.~\ref{eqn:itp} becomes:
\begin{equation}
\begin{aligned}
 i_{\psi,\omega} &= \int_{-T/2}^{T/2} \left(\int_{-\infty}^{t} a(\tau)\text{cos}(\omega t-2\omega\tau) d\tau\right) b\cos\left(\omega t-\psi\right) dt \\ &\approx \int_{-\infty}^{t} a(\tau)b\cos(\psi-2\omega\tau) d\tau = \int_{-\infty}^{t} a(\tau)f_{\psi,\omega}(\tau) d\tau.
  \label{eqn:mitp}
\end{aligned}
\end{equation}

When sinusoidal reference signals with different frequencies $\omega_l$ and phases $\psi_k$ are measured sequentially, one obtains multiple channels of raw measurements.
The multi-channel measurements at any pixel $\vec{i}_{\vec{\psi},\vec{\omega}}$ can be interpreted as a point in a multi-dimensional space, and while difficult to model analytically, there is structure in this space that can be exploited through learning.

In the ideal case of a single bounce, the set of all possible measurement vectors $\vec{i}_{\vec{\psi},\vec{\omega}}$ forms a ``single-bounce measurement manifold'' defined by Eq.~\ref{eqn:itp}.
If only one frequency is used, measurements affected by MPI lie on the same manifold, and it is therefore impossible to identify and correct them.
On the other hand, in the case of multiple frequencies, the manifold becomes a non-linear subspace, and MPI-affected vectors do not lie on it anymore.
The MPI problem can then be recast as \new{one of} mapping\old{of a set of} real measurements, possibly affected by MPI, to the ideal one-bounce manifold, which is also the idea \new{behind}\old{at the base of} many existing approaches for MPI correction.

\subsection{The Impact of Motion}
Real scenes are\old{only} rarely static.
Because of the lateral and axial motion of\old{the} objects with respect to the camera, \new{sequential} correlation measurements $\vec{i}_{\vec{\psi},\vec{\omega}}$\old{taken sequentially} are misaligned.
Moreover, the axial component of the motion also changes the scene response function $a(\tau)$: for example, even in the simple case of a single bounce, the term $\tau_0$ in Eq.~\ref{eqn:itp} changes with the axial motion of the object, whereas the measured intensity varies proportionally to the inverse-square of\old{object} distance. 
In our indoor experimental setting we found both these phenomena to contribute significantly to the depth reconstruction error.
Motion can even generate blur and Doppler within each raw correlation measurement, but we found these last effects negligible when compared to the previous ones.



\subsection{Characterization of Kinect 2}
\label{sec:kin}
The Kinect 2 is a well-documented, widely used ToF camera, with an open-source SDK (libfreenect2~\cite{libfreenect2}) that exposes raw correlation measurements and provides a baseline algorithm for benchmarking (indicated as LF2 in the following). We carefully characterize the camera functions, shot noise, vignetting, and per-pixel delay, to produce accurate simulations and mitigate the data shift between synthetic data from the FLAT dataset and real scenarios.

\begin{wrapfigure}{r}{0.37\textwidth}
    \centering
    \includegraphics[width=0.35\textwidth, clip=true, trim = 0cm 0.1cm 0.9cm 0.4cm]{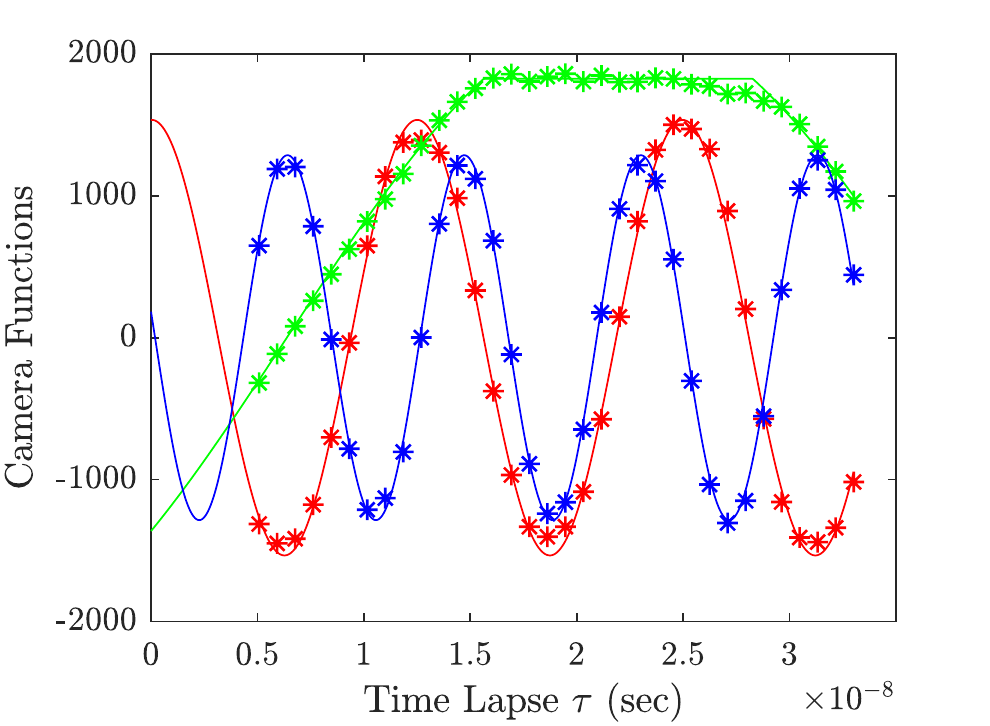}
  \caption{Calibrated camera functions for Kinect 2; the points represent experimental data, the continuous line are the fitted camera functions.}
  \label{fig:camfunc}
\end{wrapfigure}

The Kinect 2 uses three modulation frequencies, each with three phases, for a total of nine camera functions $\vec{f}_{\vec{\psi},\vec{\omega}}(\tau)$.
To calibrate the camera functions, we carefully align the optical axis to be normal to a Lambertian calibration plane.
We place a light absorption tube in front of Kinect's light source to narrow down the beam emission angle and limit MPI. 
We translate the plane to known distances $\{d_j\}_{j=1..N}$ to obtain a series of raw measurements $\{\vec{i}_{\vec{\psi},\vec{\omega}}(d_j)\}_{j=1..N}$ that approximate $(d_j)^{-2}\vec{f}_{\vec{\psi},\vec{\omega}}(2d_j/c)$ up to a constant scale, for every pixel.
After removing the squared-distance intensity decay $(d_j)^{-2}$, we have a series of observations of the camera functions $\{\vec{f}_{\vec{\psi},\vec{\omega}}(2d_j/c)\}$. Fig.~\ref{fig:camfunc} shows
three camera functions fitted by this method, parameterized as $b\cos(\psi-2\omega\tau)$ (red and blue curve in Fig.~\ref{fig:camfunc}), and $\max(\min(b_1\allowbreak\cos(\psi-2\omega\tau), b_2), -b_2)$ (green curve in Fig.~\ref{fig:camfunc}).

As for the shot noise, we assume each pixel to be independent from the others.
We acquire data from 15 scenes; for each raw correlation measurement, we compute the per-pixel expected value as the average of 100 measurements. For any expected value, we collect the empirical distribution of the noisy samples in a \new{lookup table}\old{look-up-table}, that is used to generate noisy data in simulation.   
For the complete explanation of the calibration procedure, including vignetting and per-pixel time delay, we invite the reader to refer to the Supplementary, \new{which}\old{that} also includes more experimental results.

\section{The FLAT Dataset}
\label{sec:frame}

An ideal dataset for training and validating ToF algorithms is large; allows simulating different camera response functions (like those in Section~\ref{sec:kin}); allows including MPI, shot noise and motion; and exposes raw correlation measurements.
We created the FLAT dataset with these principles in mind.

FLAT contains $2000$ scene response functions, $\{a^j(\tau,x,y)\}_{j=1.. 2000}$, where we make the dependence on pixel $(x,y)$ explicit. Each of these is computed through transient rendering~\cite{Jar14}, which simulates the irradiance received by the camera sensor at every time frame, after sending an impulse of light to the environment.
The output of the renderer is a $n_\tau\times n_x\times n_y$ tensor, i.e. a discretized version of $a^j(\tau,x,y)$.
The scenes \new{in}\old{of} the dataset are generated from 70 object setups; each setup has 1 to 5 objects with lambertian surface and uniform albedo; their 3D models \new{are}\old{come} from the Stanford 3D Scanning Repository~\cite{curless1996volumetric} and the online collection \cite{vt_collections}. 
We render each setup from approximately 30  point of views and orientations, at a spatial resolution of $(n_x = 424) \times (n_y = 512)$ pixels and for $n_\tau = 1000$ consecutive time intervals (each interval is $5e^{-11}$sec long); the horizontal field of view is 70 degrees (corresponding to the Kinect 2 camera).
Since bi-directional path tracing is used to sample and simulate each light ray, $a^j(\tau,x,y)$ does simulate MPI. 
From the discretized version of $a^j(\tau,x,y)$, any raw measurement $i_{\psi,\omega}$ can be obtained as in Eq.~\ref{eqn:mitp}, for any user-provided camera function $f_{\psi,\omega}(t)$ (like, for instance, the ones we estimated for Kinect 2).

The FLAT dataset offers the possibility to augment the raw measurement with shot noise, textures, vignetting, and motion.
Within FLAT, we provide the code to apply any vignetting function and shot noise coherently with the simulated camera, while MPI and camera functions are handled as described in the previous paragraph.
\new{As a consequence,}\old{A consequence of this is that} a physically correct simulation of different camera functions, MPI, vignetting, and shot noise is a computationally light task within FLAT.
On the other hand, texture and motion are more expensive to render exactly.
Since each scene in the FLAT dataset takes on the average 10 CPU hours to render, creating a large set of scenes with different textures and motions would require tens to hundreds of times \new{longer}\old{more time}.
We handle this by providing tools to approximate texture and motion in real time (as specialized forms of data augmentation), while still providing a small testing set within FLAT with exact motion, texture, and\old{exact} rendering. 

We approximate textures on the training data, by pixel-by-pixel multiplication of the rendered $\vec{i}_{\vec{\psi},\vec{\omega}}$ with texture maps from the CURET texture dataset~\cite{Dan1999}.
This is an\old{crude} approximation that ignores the recursive impact that real textures have on MPI, but we have found that it is nonetheless useful as a form of data augmentation.

The FLAT dataset offers two different methods to augment the simulated raw measurements with approximate motion.
To illustrate the first one, let us consider the Kinect 2, where nine correlation measurements of a static scene, $\vec{i}_{\vec{\psi},\vec{\omega}}=(i_{\psi_1,\omega_1},\ldots,i_{\psi_9,\omega_9})$ are simulated.
We generate a random 2D affine transform $T$ and apply it to create a smooth motion as $i_{\psi_j,\omega_j}'(x,y) = i_{\psi_j,\omega_j}(T^{j-5}(x,y))$\new{, where $T^{n}(x,y)$ is transforming $(x,y)$ by $T$ for $n$ times, and $T^{-n}(T^{n}(x,y)) = (x,y)$}. Notice that the first and last measurement will achieve the largest movement.
To obtain a more complex movement, we simulate the motion of two or more objects with different affine transforms and composite the scene based on their depths.
This approximate motion model is fast, but does not reproduce the MPI interaction between the objects in the scene.
The second motion approximation method 
takes in input a rendered scene response function, $a(\tau,x,y)$.
We generate then a random, 3D affine transformation and apply the corresponding displacement $(v_x,v_y,v_z)$ to it, i.e., $a'(\tau,x,y)=a(\tau+v_z/c,x+v_x,y+v_y)$. Then we use Eq.~\ref{eqn:mitp} to compute one of the nine raw measurements. 
As in the previous method, the transform is applied multiple times to create a smooth motion between the nine measurements.
This method is computationally more expensive compared to the previous one, but physically more accurate.

\begin{figure}
\begin{minipage}{\linewidth}
\scriptsize
\begin{center}
\begin{tabular}{|c|c|c|c|}
\hline
  Name & MOM + LF2 & MOM-MRM + LF$2^\ast$ & MRM + LF$2^\ast$ \\
\hline
 Training Data & Motion, MPI & Motion, MPI & Static, MPI \\
\hline
 \multirow{2}{*}{Architecture} & \multirow{2}{*}{Encoder-Decoder} & Encoder-Decoder- & \multirow{2}{*}{KPN($1\times1\times9$)} \\
 & & KPN($3\times3\times1$) &  \\
 \hline
 Depth Reconstruction & LF2 & LF2, no filter & LF2, no filter \\
 \hline
 Training Loss & L2, Velocity & L2, Raw Intensity & L2, Raw Intensity \\
 \hline
 Fine Tuning Loss &  & L2, Depth & L2, Depth \\
 \hline
\end{tabular}
\captionof{table}{Training specification.
}
\label{tab:trainspec}
\end{center}
\end{minipage}
\begin{minipage}{\linewidth}
\centering
\subfloat[True depth]{\includegraphics[height=0.15\textwidth]{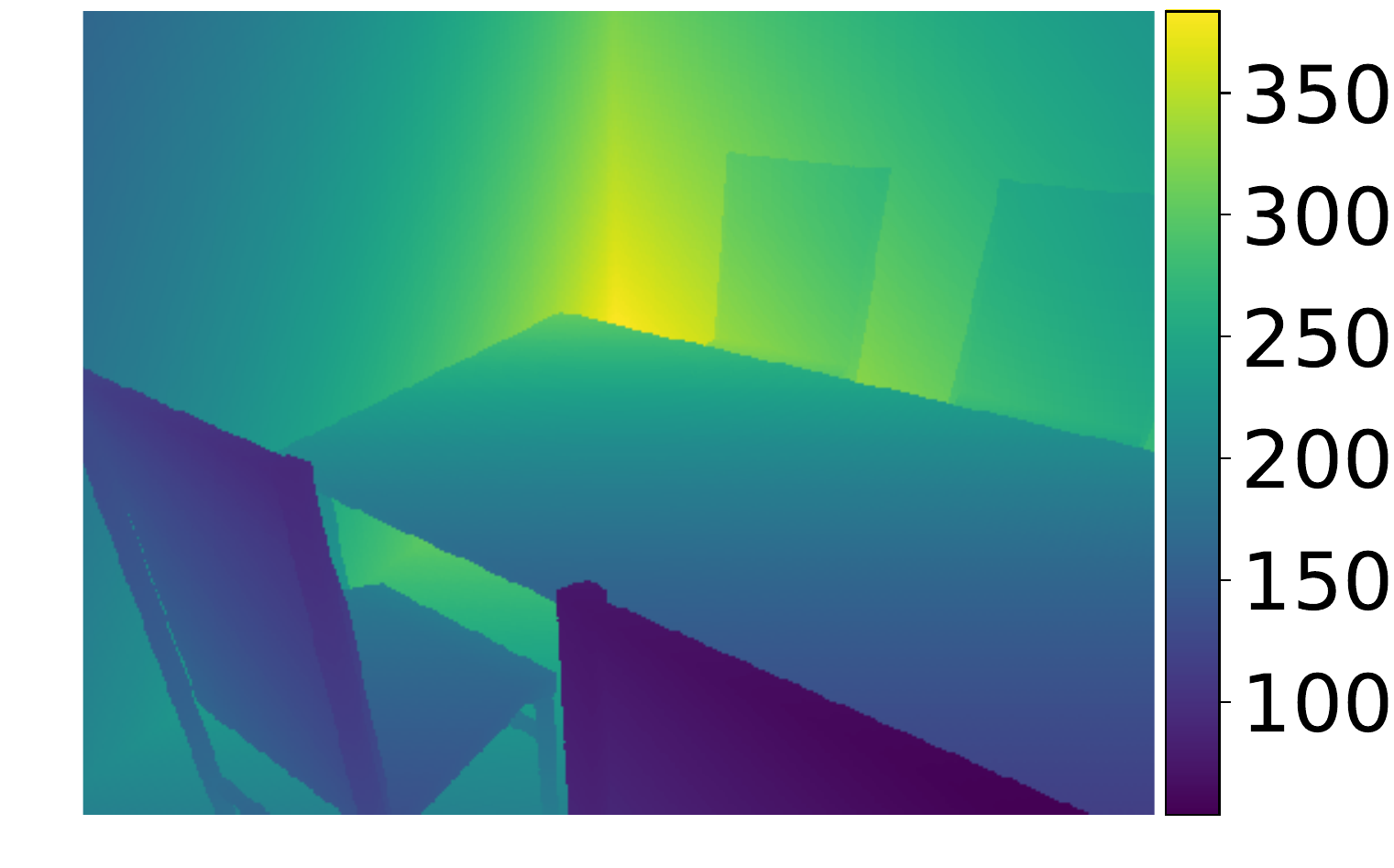}}
\subfloat[Motion]{\includegraphics[height=0.15\textwidth]{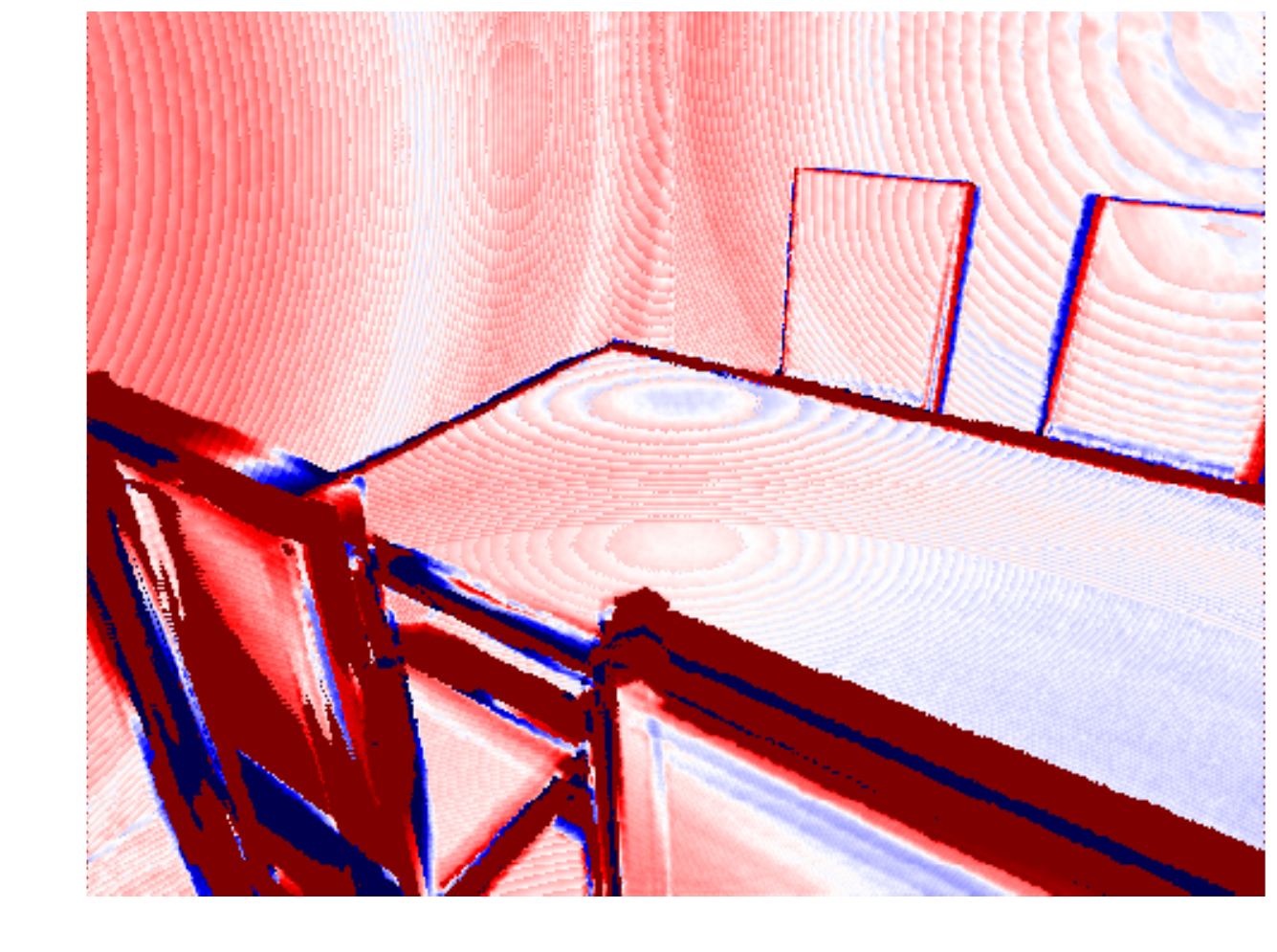}}
\subfloat[Shot noise]{\includegraphics[height=0.15\textwidth]{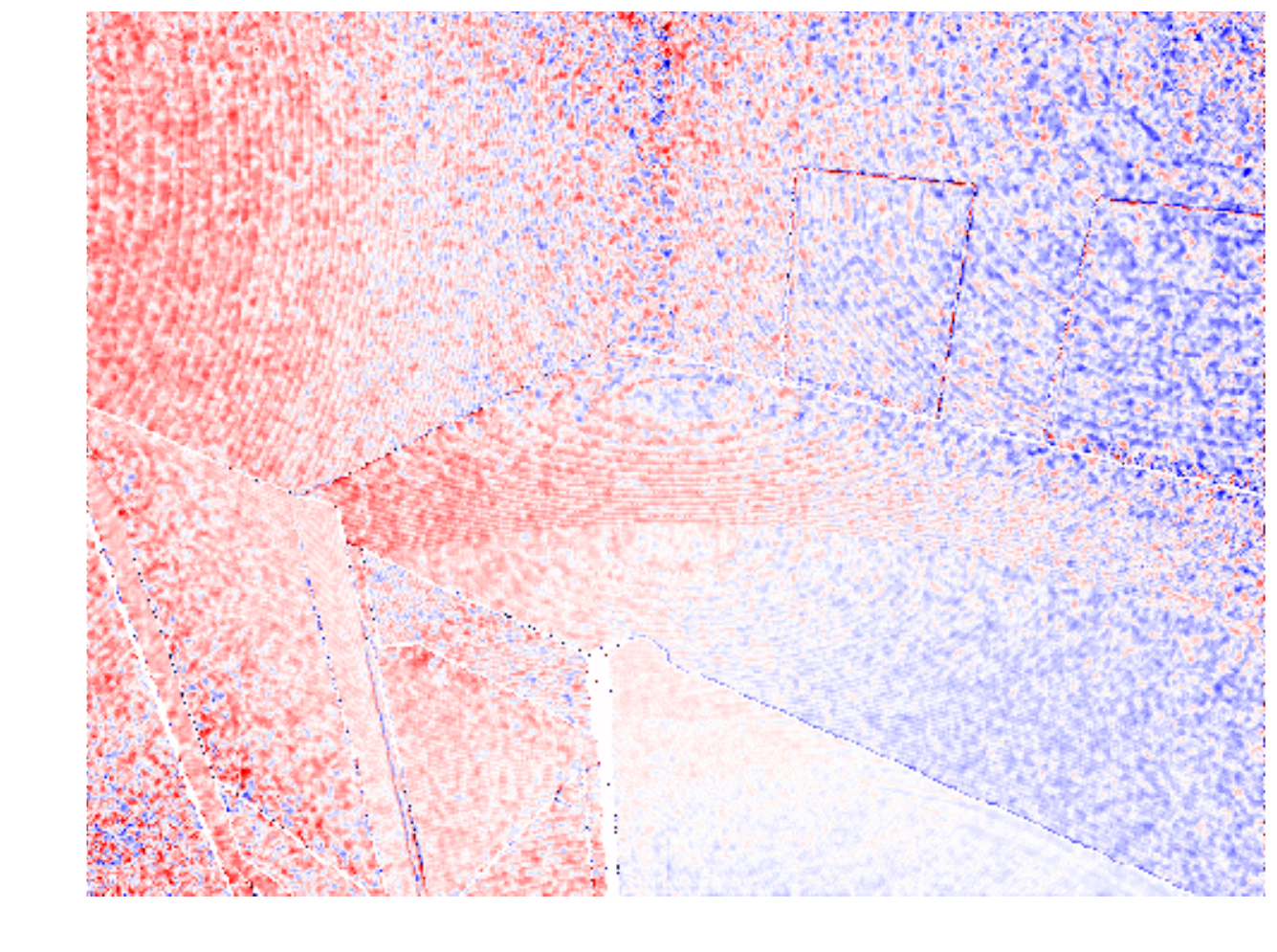}}
\subfloat[MPI]{\includegraphics[height=0.153\textwidth]{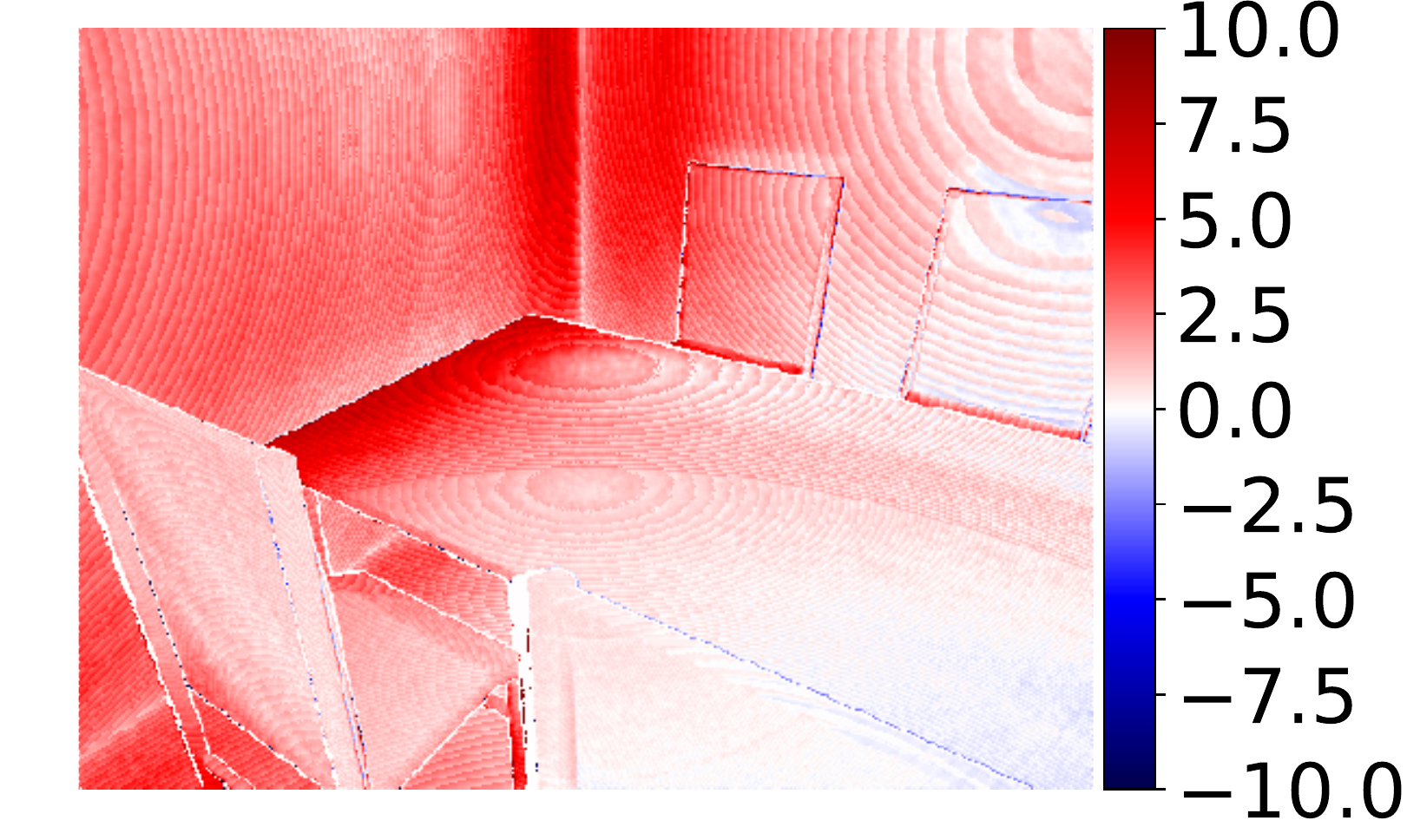}}
\captionof{figure}{Effect of non-idealities on the LF2 depth reconstruction error, in cm.}
\label{fig:noises}
\end{minipage}
\end{figure}

\section{Network Architecture}

We propose a two-module DNN (Fig.~\ref{fig:teaser}), to deal with MPI, motion, and shot noise in the domain of raw correlation measurements. We demonstrate the use of the modules on \new{data from the} Kinect 2.
The two modules can be integrated into the LF2~\cite{libfreenect2} reconstruction pipeline, which is basically an implementation of Eq.~\ref{eq:gs} on Kinect 2.
We leverage the differentiability of Eq.~\ref{eq:gs} to exploit, among other forms of training for the DNN, training using a loss function in the depth domain.

The first module (MOtion Module, MOM) is an encoder-decoder inspired by Flownet~\cite{Dos15}.
The aim of MOM is to output a velocity map to align the nine raw channels measured by Kinect2.
Differently from the original design of Flownet, which computes the optical flow between two images in the same domain, MOM aligns raw correlation measurements taken with different camera functions, and therefore correlated, but visually different.
Moreover, MOM takes in input nine misaligned channels and outputs eight optical flows at the same time, while Flownet deals with only one pair of images and one optical flow.
The second module (Multi-Reflection Module, MRM) is based on a kernel-predicting network (KPN), that has been  effectively used for burst denoising on shot noise~\cite{mildenhall2017burst}.
MRM outputs nine spatially varying kernels for every pixel; each kernel is locally convolved with the input raw measurements, to produce a clean raw measurement by jointly removing shot and MPI noise on every channel.

Table~\ref{tab:trainspec} shows the details of different variations of the basic DNN architecture that we consider in our experiments.
Notice that, when we use the MRM module, we modify the LF2 pipeline to remove bilateral filtering from it,\old{the rationale being that} \new{because} denoising is already\old{(and possibly better)} performed by MRM; we indicate this variation of the LF2 pipeline as LF2$^\ast$.
The MOM-MRM (followed by LF2$^\ast$) network inherits the encoder-decoder from MOM; training of MRM is performed with the output of MOM as input, the weights of MOM being fixed.
We start training using the L2 error of the raw correlation measurements as loss.
Then, we fine tune MRM using the L2 depth loss propagated through LF2$^\ast$.
Motion in the training data is generated using the first approximation method in the FLAT dataset.
We also tried fine tuning using the second approximation, which is physically more accurate, obtaining similar results.

\section{Experiments}
\label{sec:experiments}

To better illustrate the typical distribution of the artifacts introduced by motion, shot noise, and MPI,
we show in Fig. \ref{fig:noises} a scene from the FLAT dataset, rendered with motion, shot noise, or MPI only, and then reconstructed by the LF2 pipeline.  
Over/under shootings can be observed at the border of moving 
\begin{wrapfigure}{r}{0.6\textwidth}
\begin{minipage}[tu]{0.59\textwidth}
\centering
\includegraphics[width=0.49\linewidth, clip=true, trim=0cm 0cm 1.5cm 1cm]{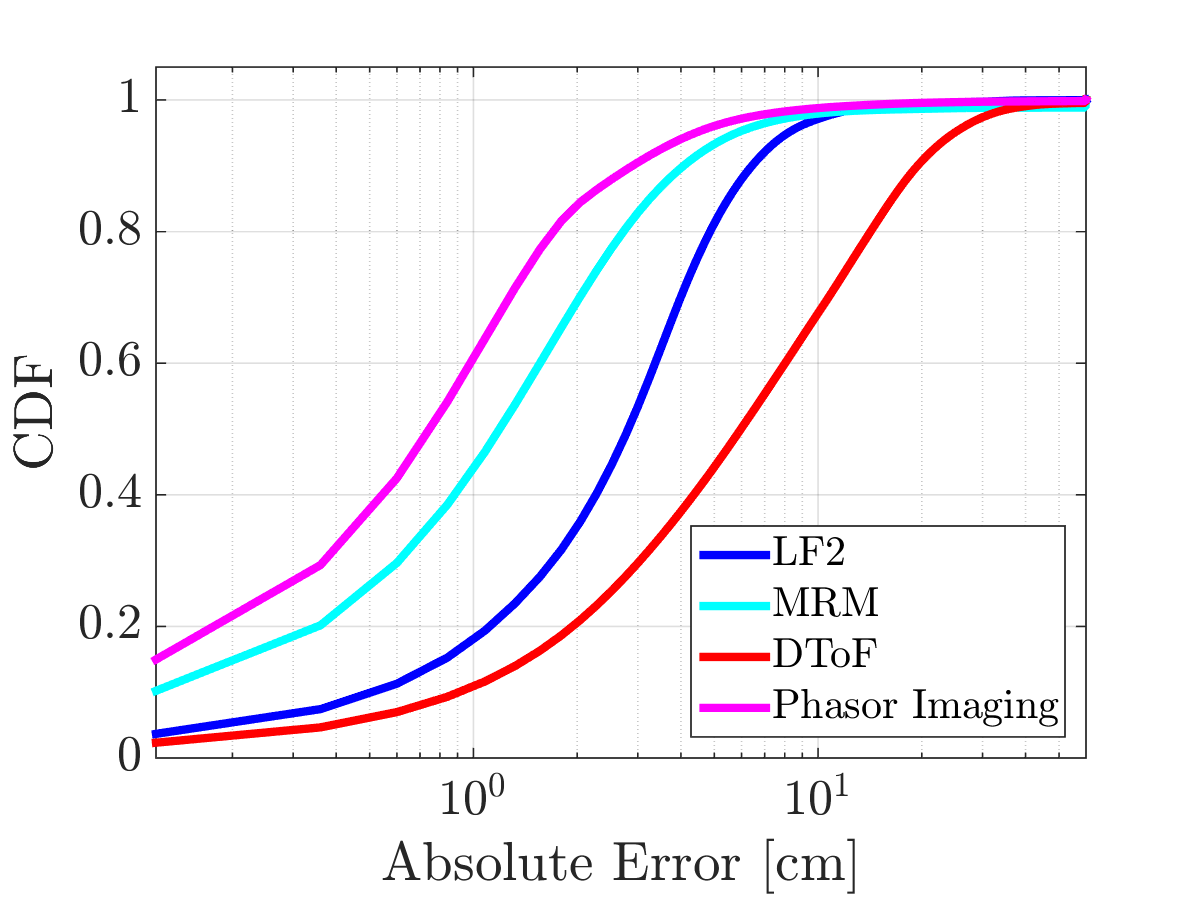}
\includegraphics[width=0.49\linewidth, clip=true, trim=0cm 0cm 1.5cm 1cm]{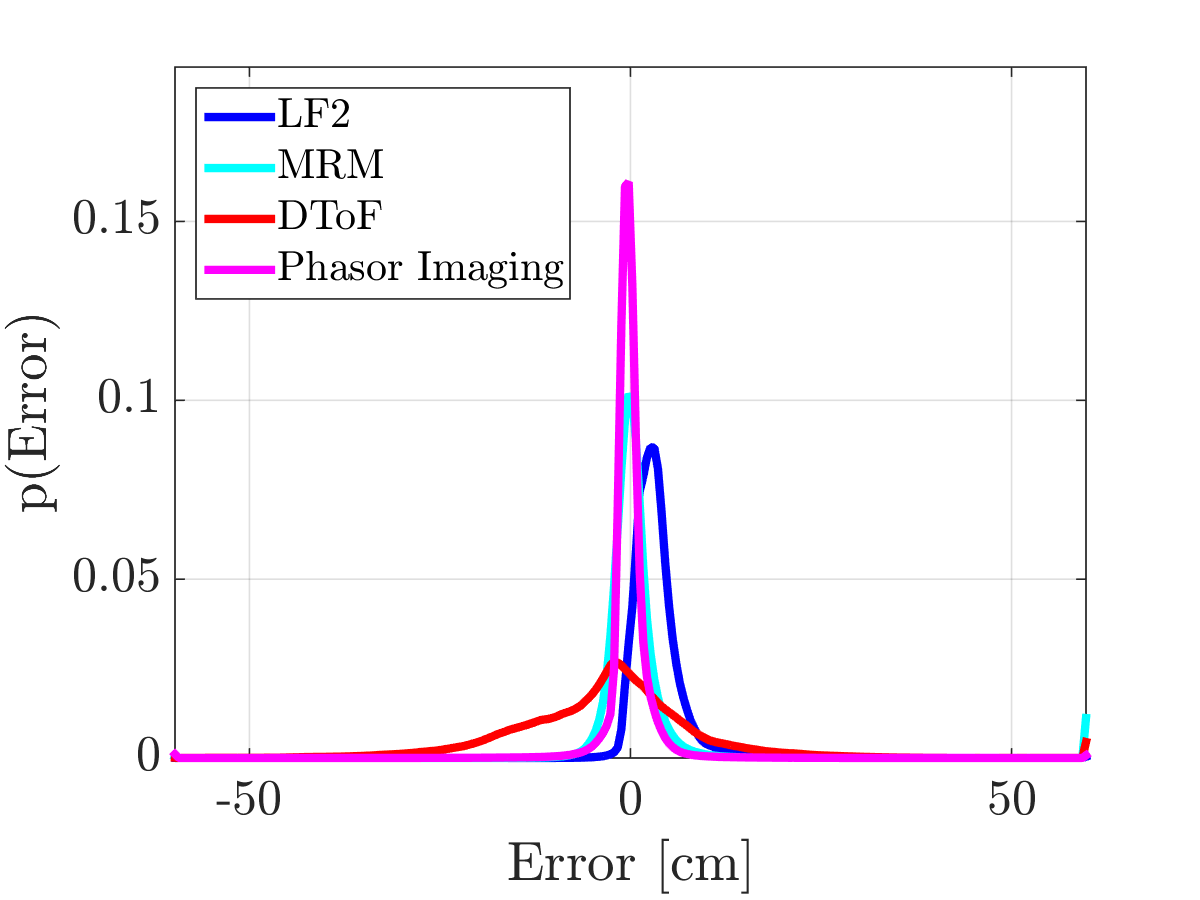}
\end{minipage}
\\
\begin{minipage}[tu]{0.5\textwidth}
\centering
\footnotesize
\begin{tabular}{|c|cc|c|}
\hline
 & Median & IQR & Prctile (90$^{th}$) \\
 & \multicolumn{2}{c|}{(error)} & (absolute error) \\
\hline
LF2 			& 2.92 (1.62\%)  & 3.01 (2.34\%)  & 6.55 (5.36\%)\\
MRM 	& -0.01 (0.00\%) & 2.63 (1.70\%)  & 4.19 (2.52\%)\\
DToF 			& -2.48 (1.80\%) & 12.36 (7.13\%) & 19.56 (9.84\%)\\
Phasor 	& -0.29 (0.12\%) & 1.62 (0.71\%)  & 1.83 (1.16\%)\\
\hline
\end{tabular}
\end{minipage}
\caption{The upper left panel shows the CDF of the depth error for LF2, MRM, DToF, and Phasor, for simulated data from  FLAT, affected by shot noise and MPI. The upper right panel shows the histogram of the error.  All pixels whose real depth is in the [1.5m, 5m] range have been considered. Our MRM outperforms LF2 and DToF, and it comes much closer to the accuracy achieved using the MPI-dedicated, higher-frequency modulations of Phasor. The table shows corresponding median and Inter Quartile Range (IQR) of the depth error, and $90^{th}$ percentile of the absolute error, in cm. The numbers in the brackets indicate the relative errors.}
\label{fig:simulated_mpi_shot}
\end{wrapfigure}
objects, where raw measurements from the foreground and the background mix due to motion.
Shot noise in $\vec{i}_{\vec{\psi},\vec{\omega}}$ creates random noise in the  depth map, especially in dark regions like background and image borders.
MPI generates a low frequency noise in areas affected by light reflection, like the wall corner in Fig. \ref{fig:noises}.  

\subsection{MPI Correction}


We first measure the effect of the MRM module on static scenes affected by MPI in the FLAT dataset, and compare it to LF2~\cite{libfreenect2}, DToF~\cite{marco2017deeptof}, and Phasor~\cite{Gup15}, that are based respectively on multi-frequency, deep learning, and custom hardware.
LF2 implements Eq.~\ref{eq:gs} on Kinect 2 and it constitutes our baseline to evaluate the improvement provided by our DNN on the same platform.
DToF and Phasor require different sensor platforms than Kinect 2, but thanks to the flexibility of \new{the} FLAT dataset, we can simulate raw measurements using their specific modulation frequency and phase, and add the same level of noise for testing.
As DToF and Phasor do not mask unreliable output pixels (like LF2 does), and Phasor's working range is limited to [$1.5$m,~$5$m], we compare the depth error for all those pixels in this range only.
Furthermore, as Phasor does not deal with shot noise, we apply a bilateral filter to remove noise from its output depth. 

\begin{figure}[t]
\centering
\subfloat[True]{
\begin{tabular}{c}
     \includegraphics[height=0.13\textwidth]{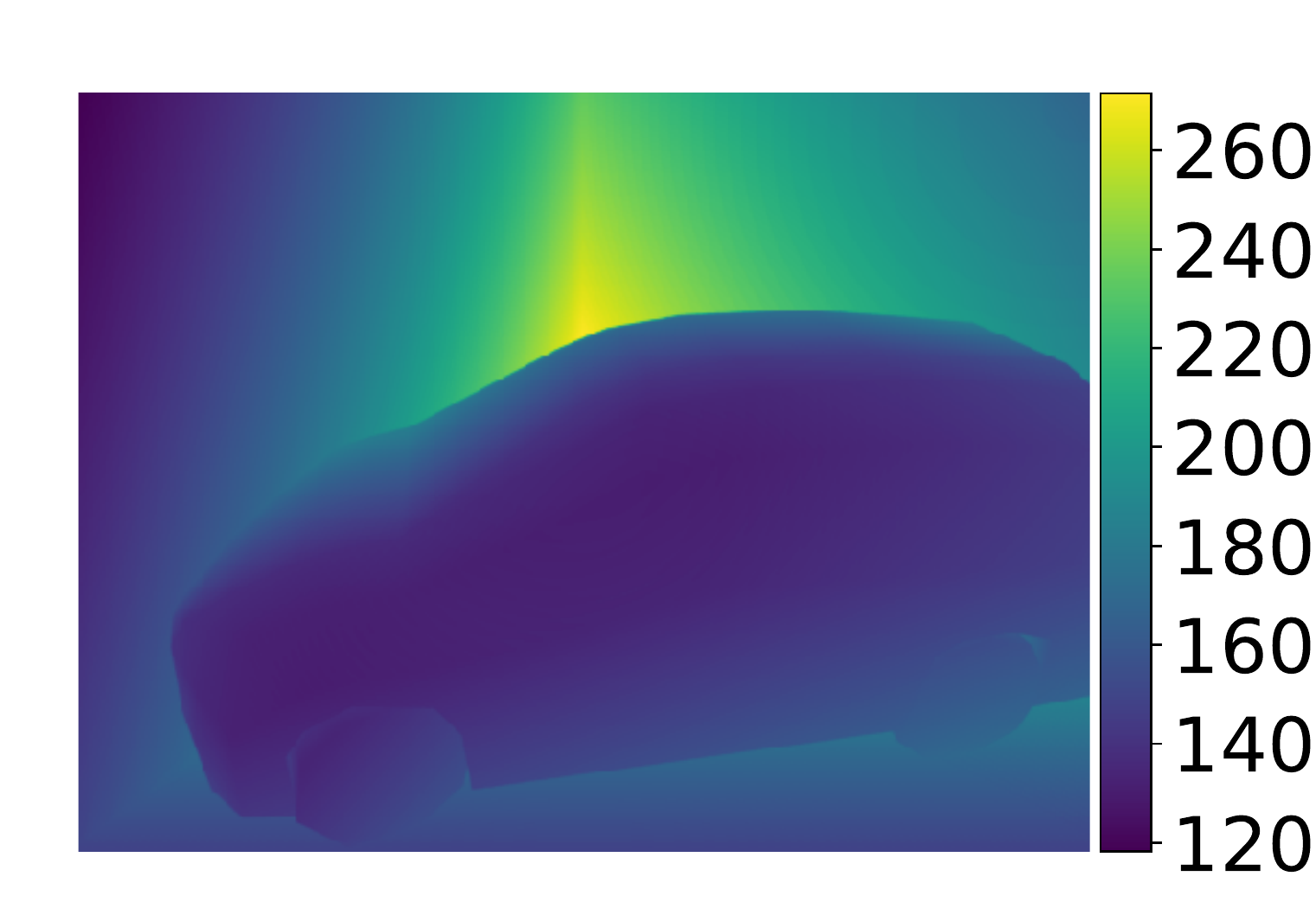} \\
     \includegraphics[height=0.13\textwidth]{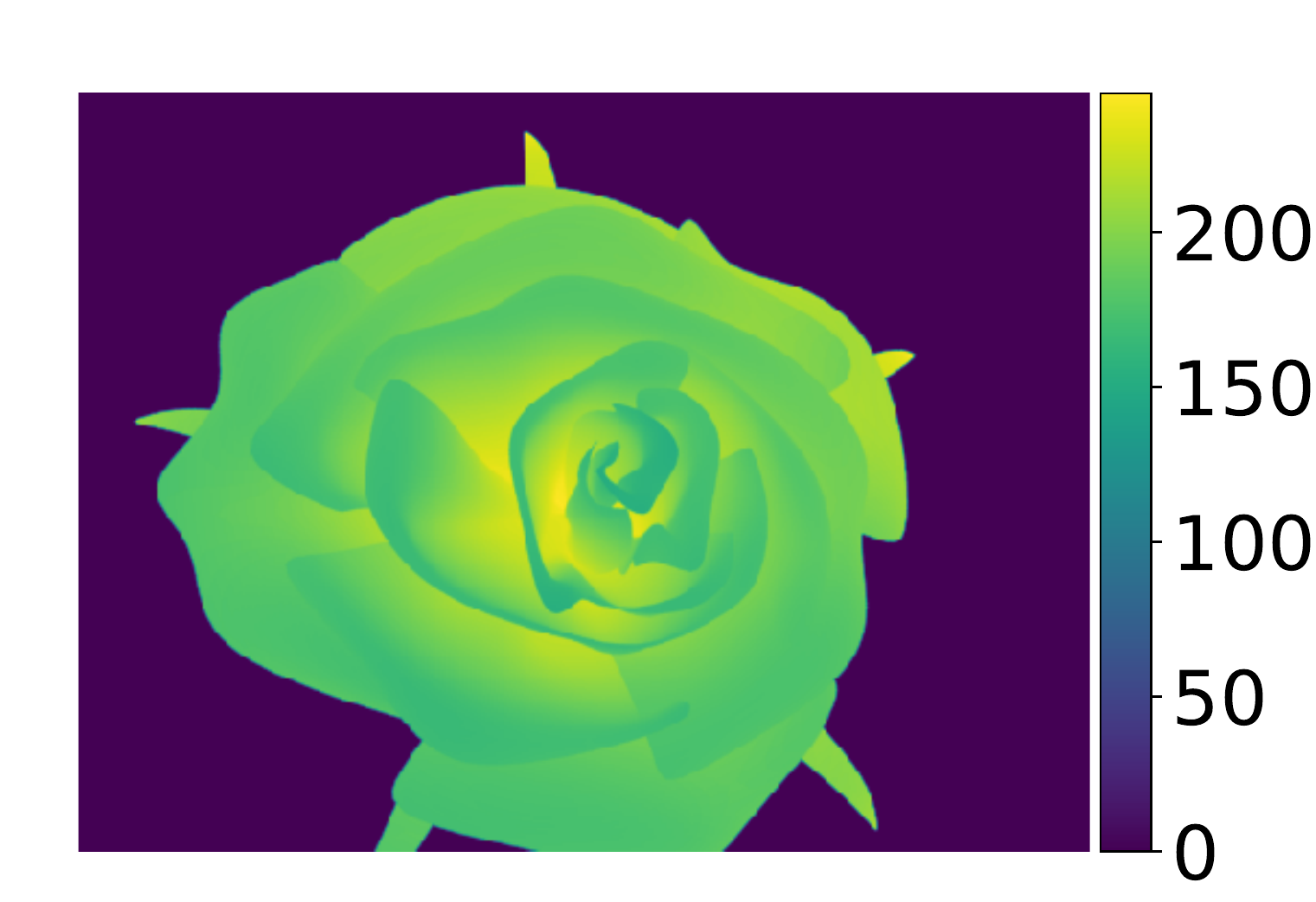}
\end{tabular}
}
\subfloat[DToF]{
\begin{tabular}{c}
     \includegraphics[height=0.13\textwidth]{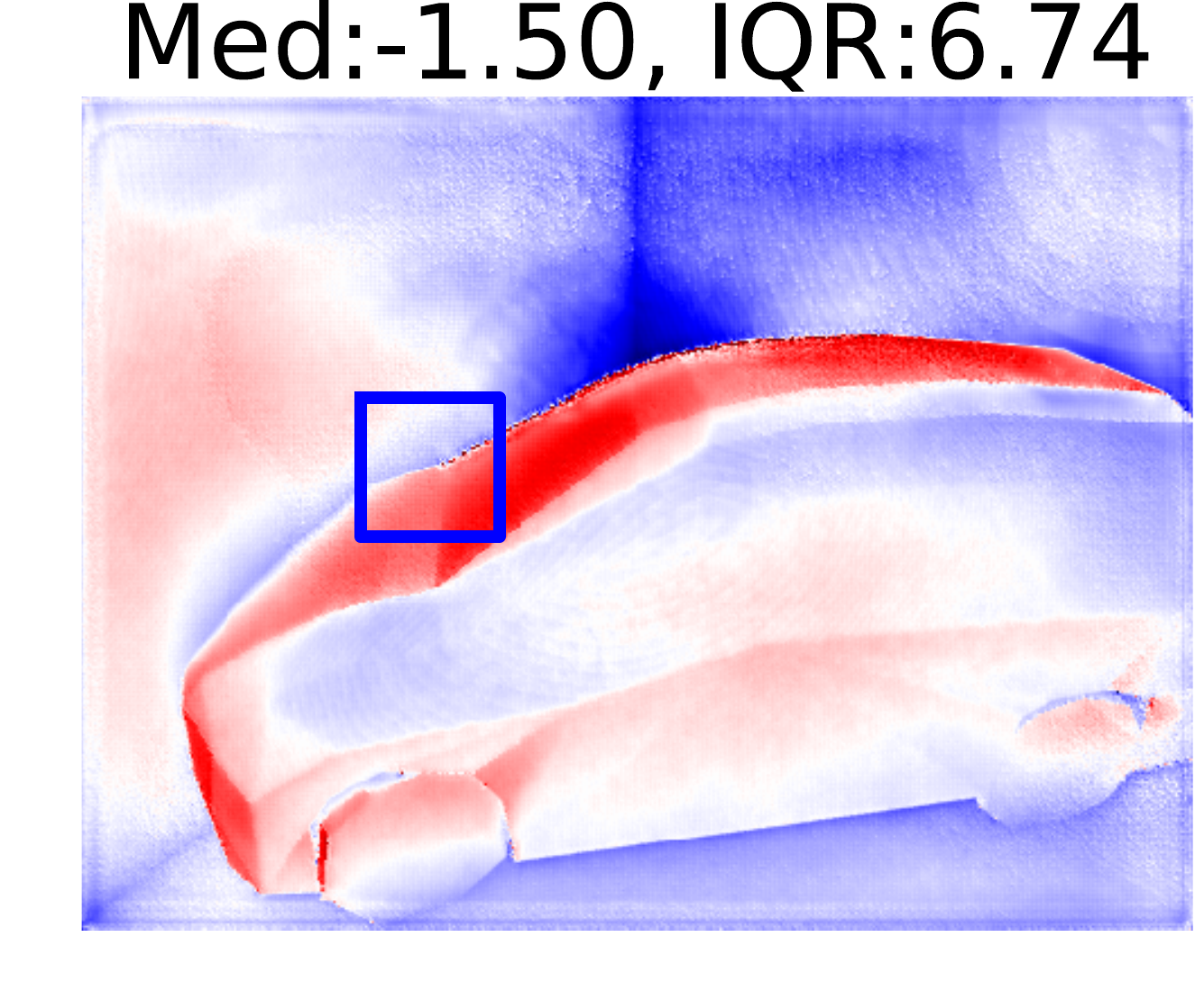} \\
     \includegraphics[height=0.13\textwidth]{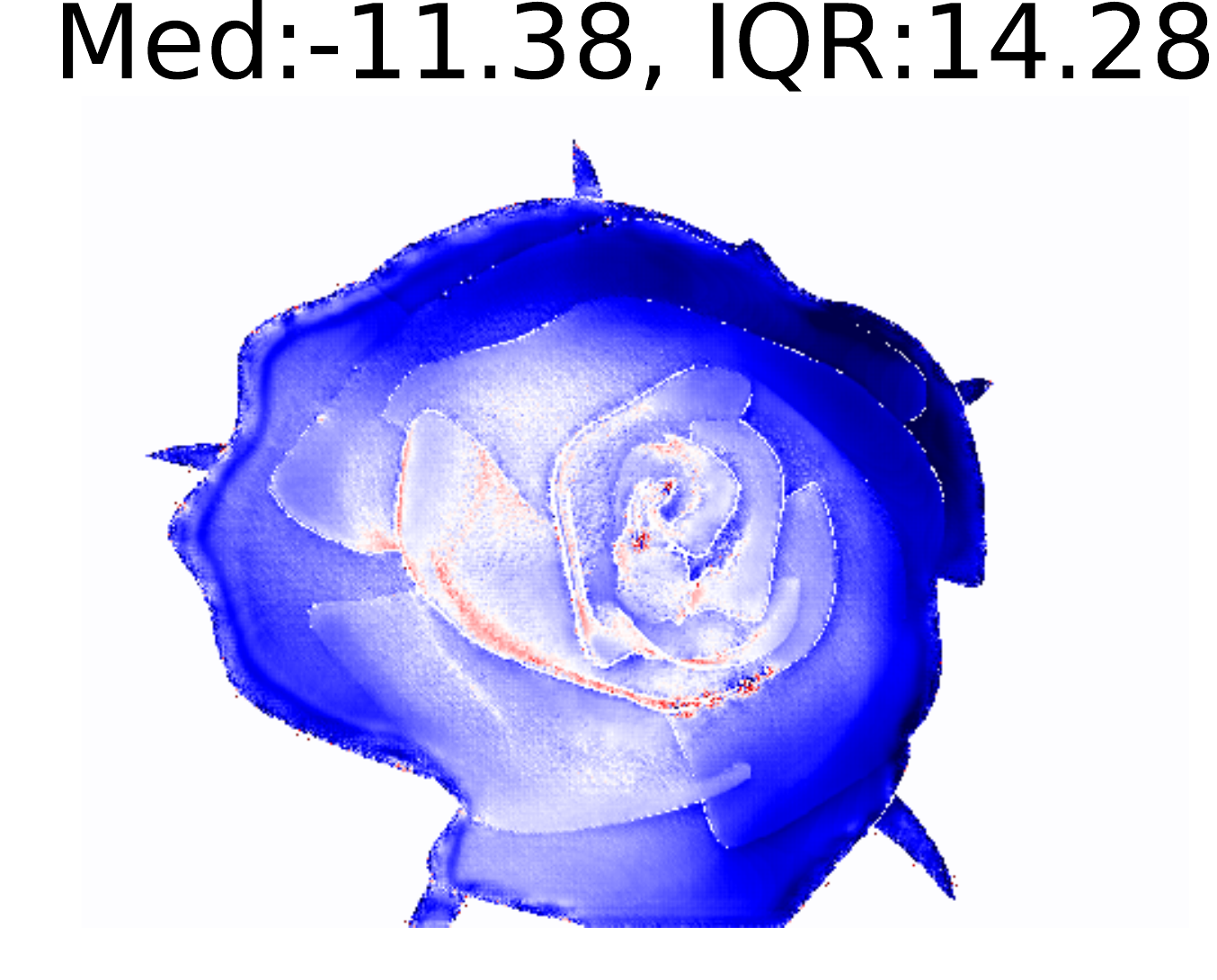}
\end{tabular}
}
\subfloat[Phasor]{
\begin{tabular}{c}
     \includegraphics[height=0.13\textwidth]{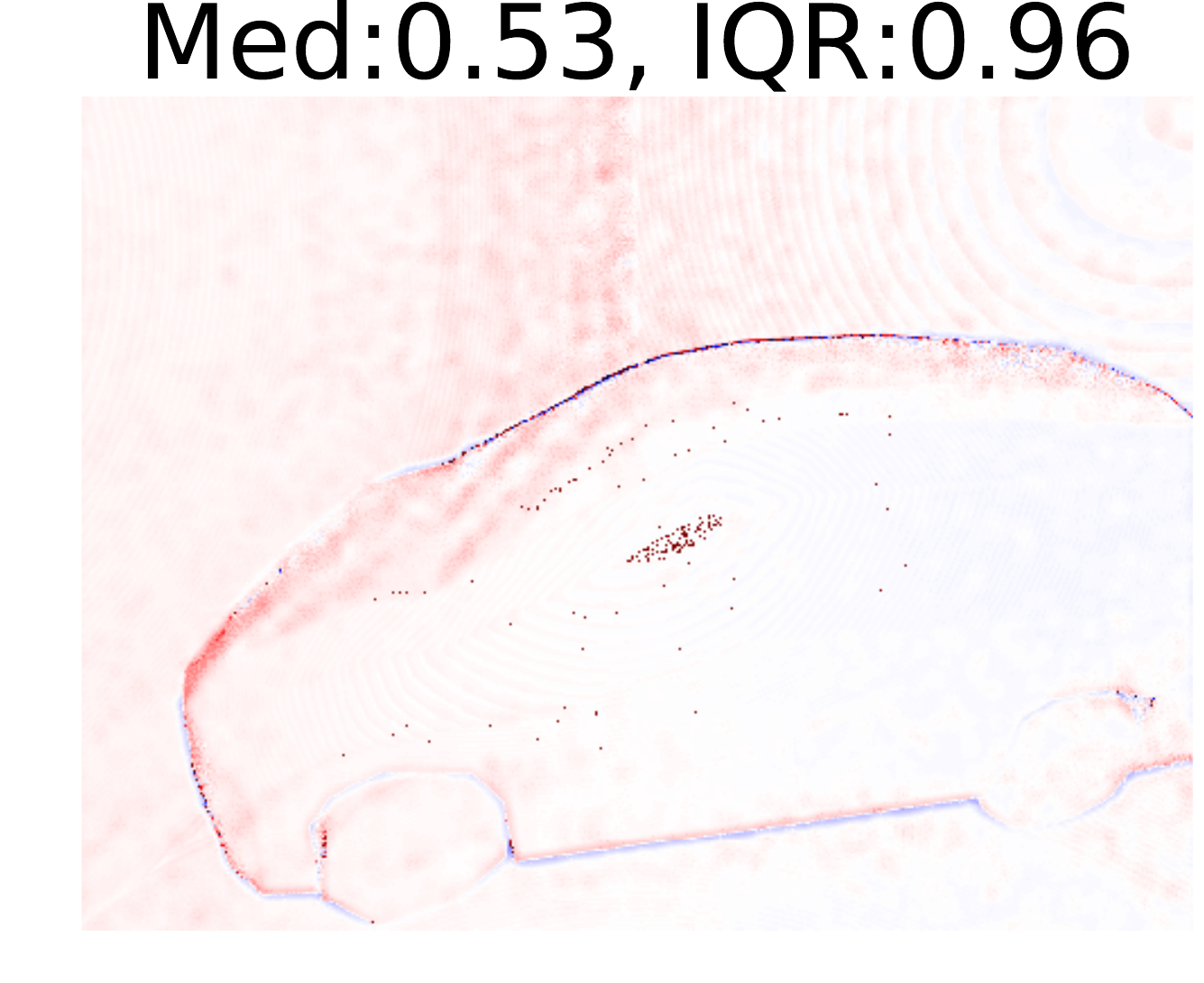} \\
     \includegraphics[height=0.13\textwidth]{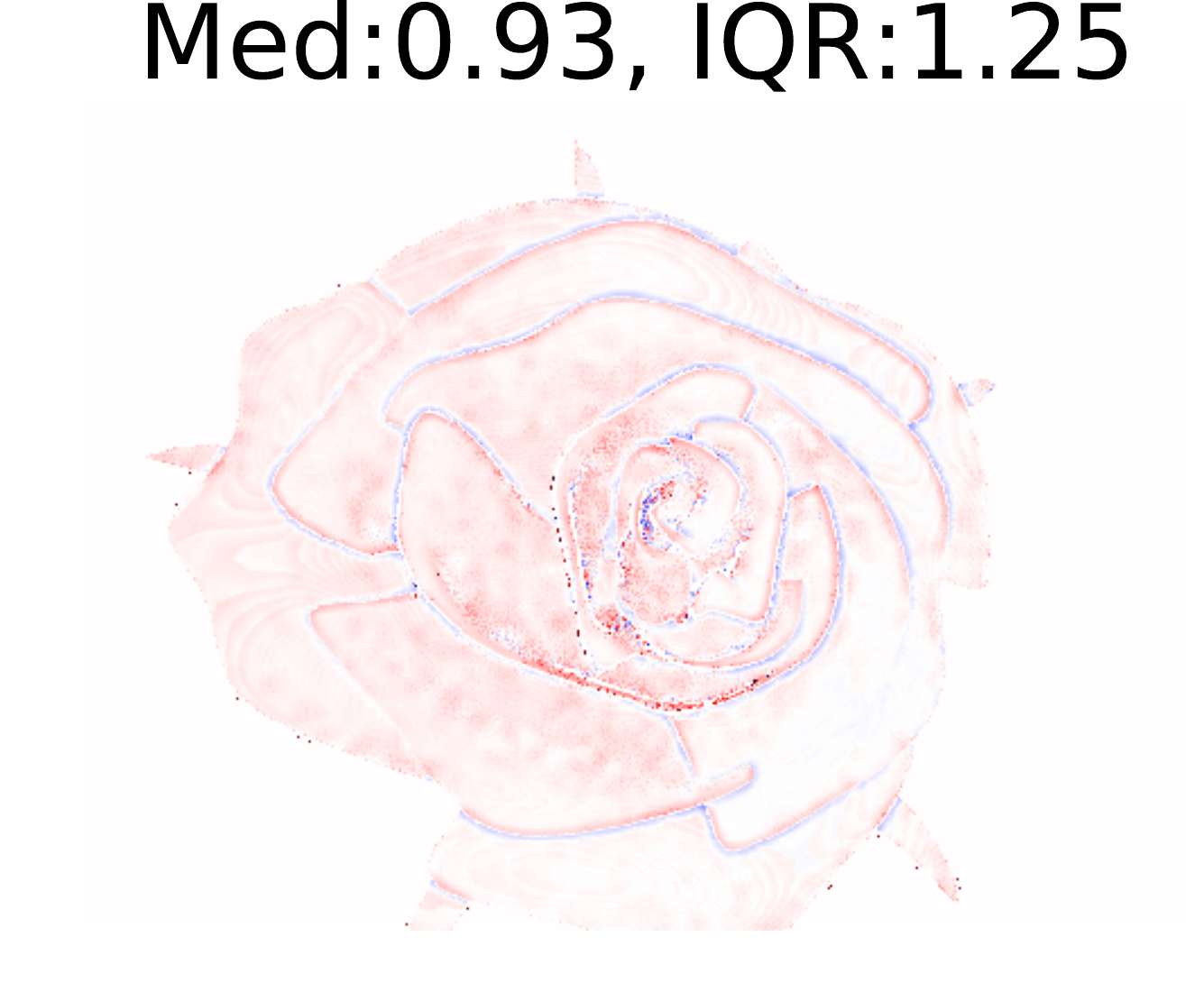}
\end{tabular}
}
\subfloat[LF2]{
\begin{tabular}{c}
     \includegraphics[height=0.13\textwidth]{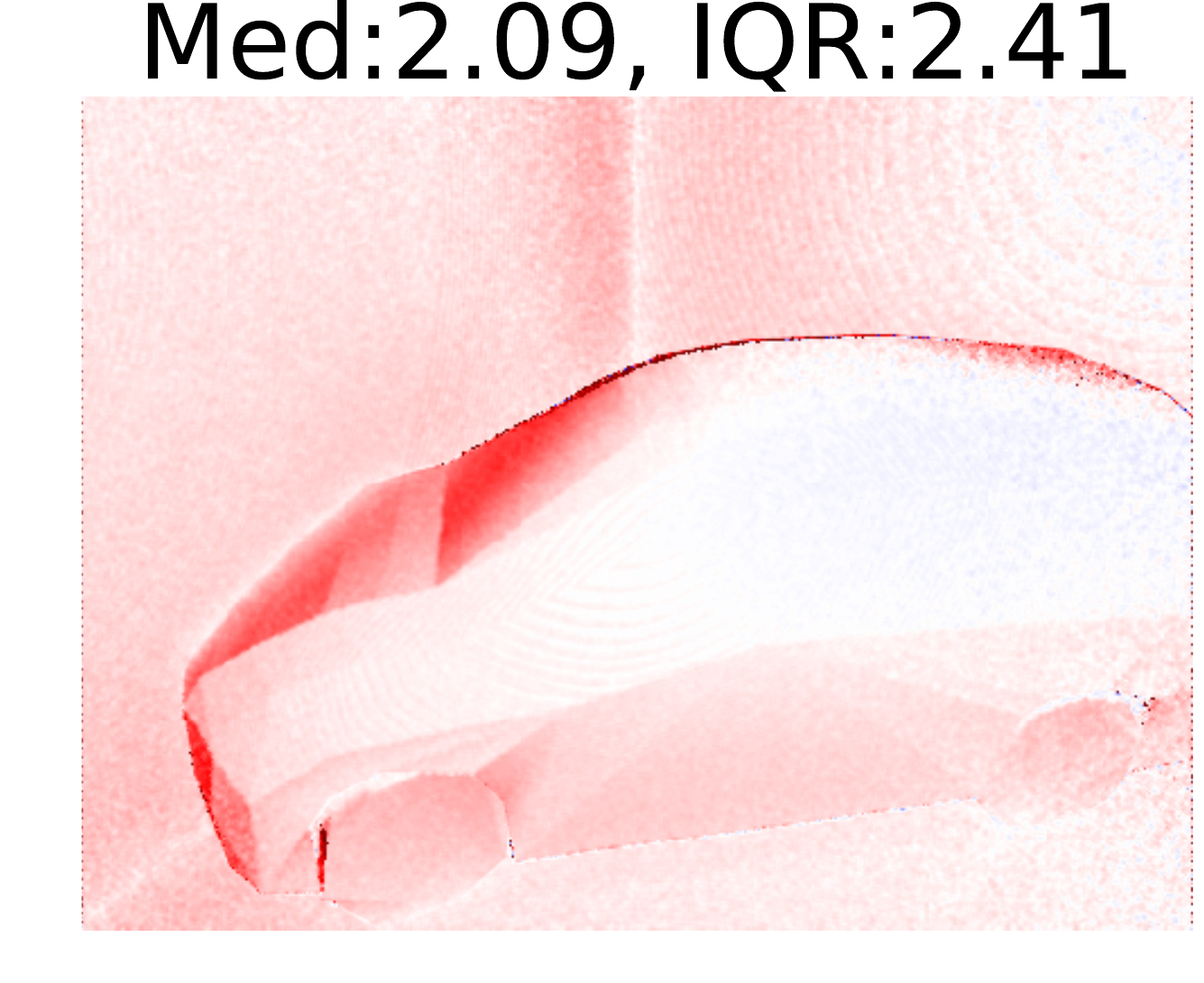} \\
     \includegraphics[height=0.13\textwidth]{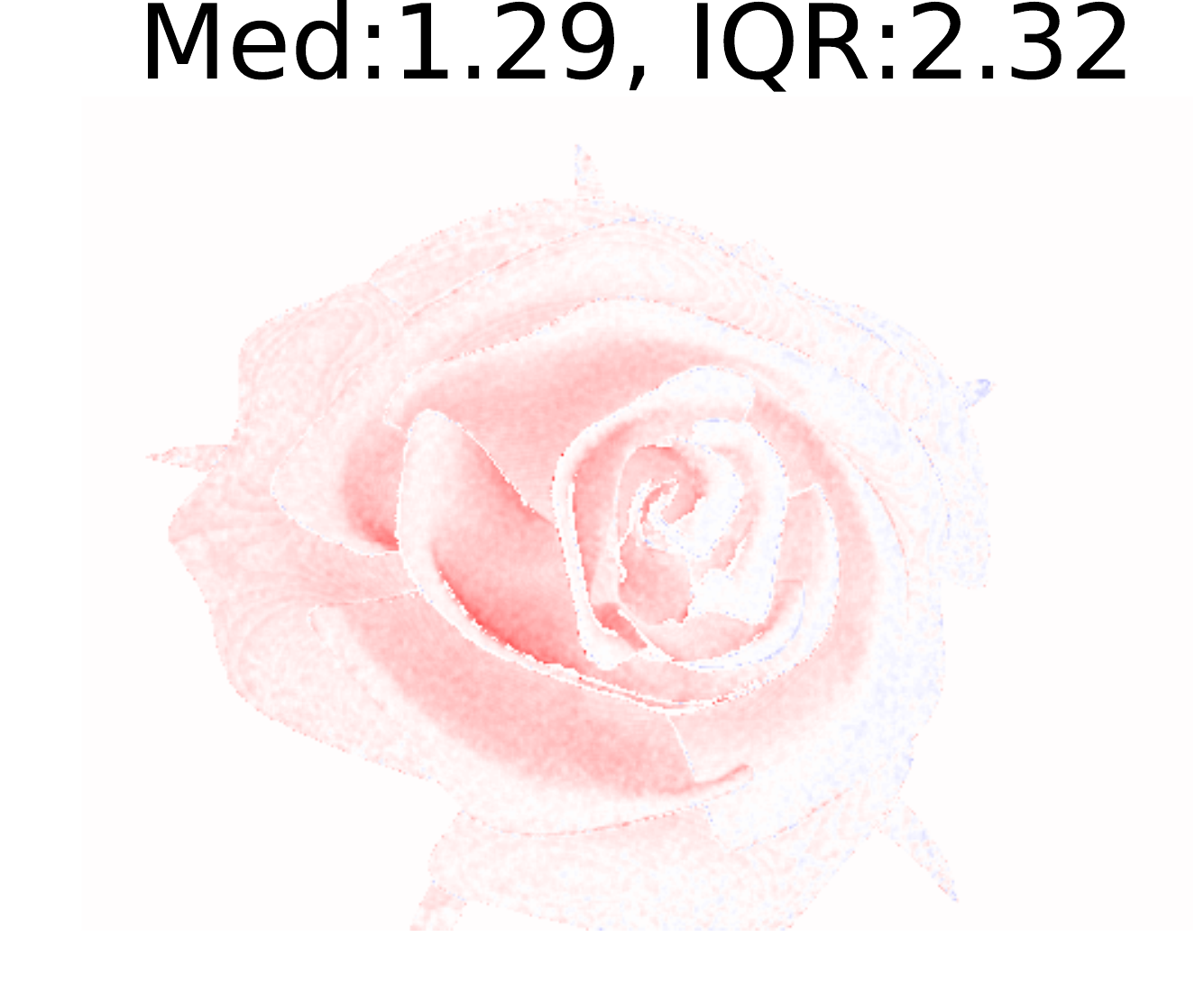}
\end{tabular}
}
\subfloat[MRM]{
\begin{tabular}{c}
     \includegraphics[height=0.13\textwidth]{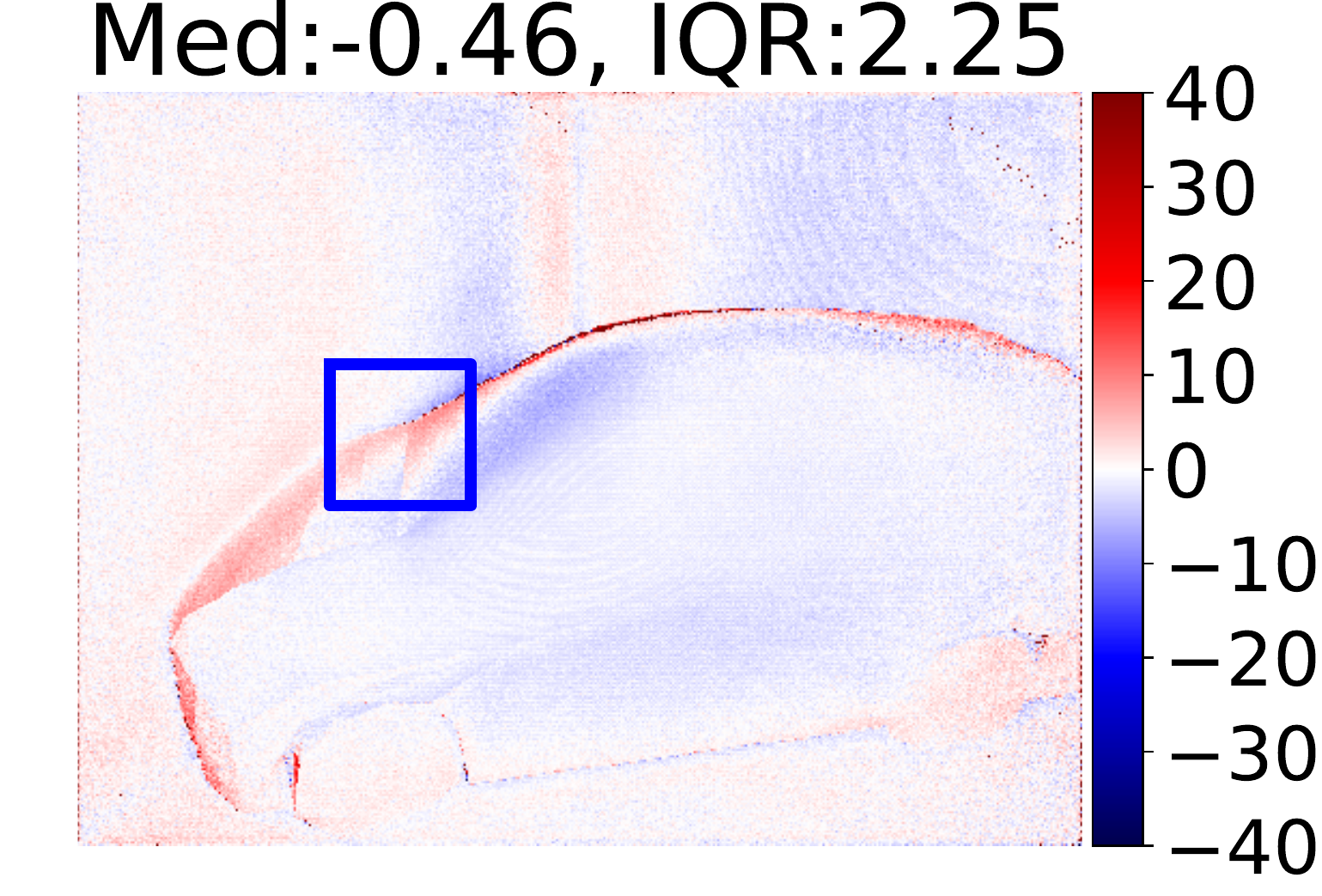} \\
     \includegraphics[height=0.13\textwidth]{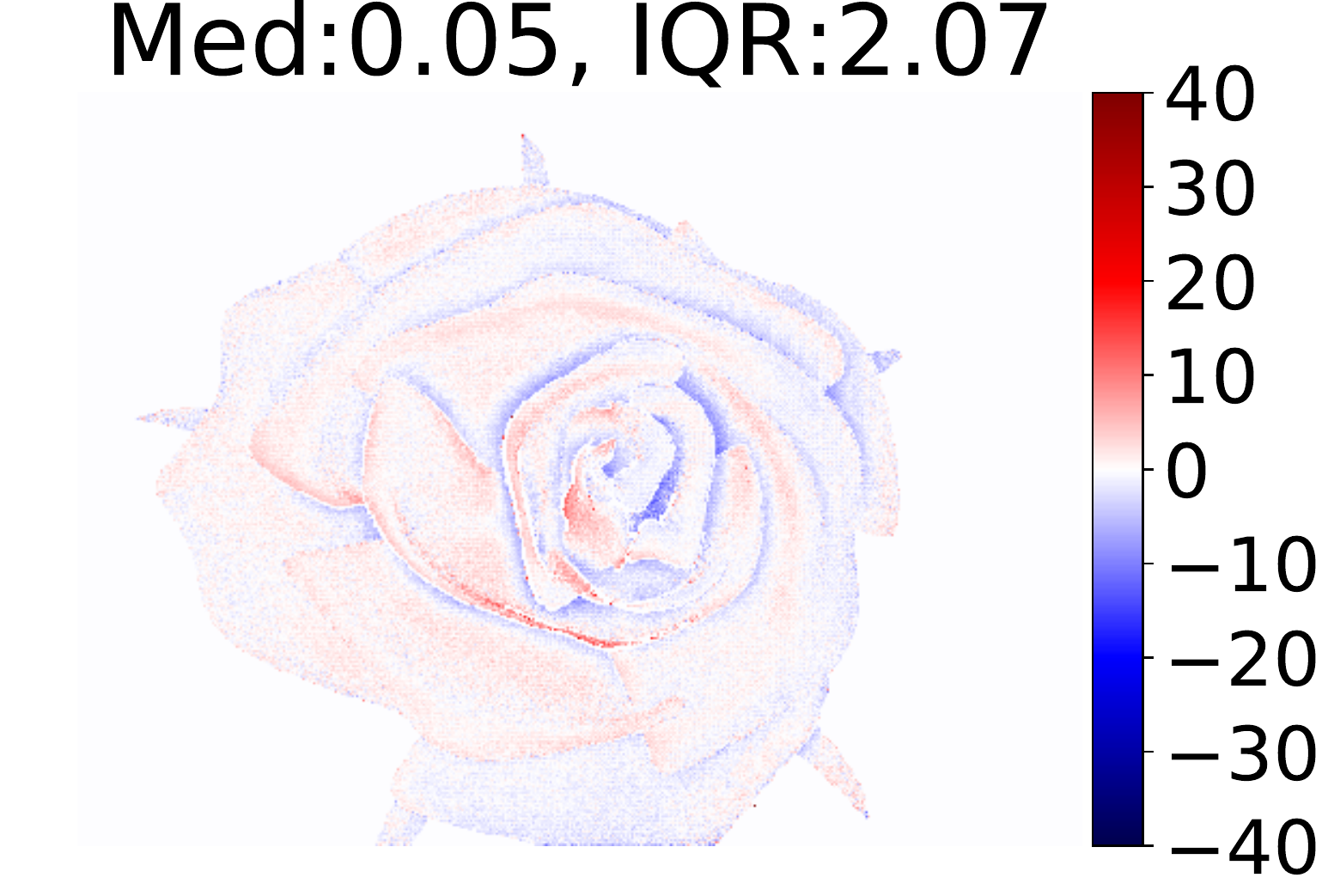}
\end{tabular}
}
\caption{Depth error for scenes from the FLAT dataset, corrupted by shot noise and MPI and reconstructed by DToF~\cite{Jar17}, Phasor Imaging~\cite{Gup15}, LF2~\cite{libfreenect2} and MRM, in cm. Errors are computed only in the unanbiguous reconstruction range of Phasor Imaging, [1.5m, 5m]; no mask is used to remove unreliable pixels. The {\color{Blue}blue} boxes in the first row show the receptive field for DToF and MRM.
}
\label{fig:errorMPISimulated_example}
\end{figure}

Results reported in 
Fig.~\ref{fig:simulated_mpi_shot} show that DToF produces the less accurate depth map. 
The median error of LF2 is biased because of the presence of MPI in the raw data---in fact, the LF2 pipeline does not include any mechanism to correct such non-ideality.
This bias is effectively removed by MRM.
\new{Our method approaches the Phasor's accuracy, without requiring expensive hardware to create very high modulation frequencies (1063.3MHz and 1034.1MHz): MRM works with Kinect 2, which uses frequencies below 50MHz.}
\old{Our method approaches the Phasor's accuracy, which however requires expensive hardware to create very high modulation frequencies (1063.3MHz and 1034.1MHz), whereas MRM is paired with Kinect 2 that uses frequencies below 50MHz.}
Fig.~\ref{fig:errorMPISimulated_example} shows the results of typical scenes from the FLAT dataset, where Phasor and MRM outperform other methods in removing MPI.
It is worth noticing that high frequency modulation signals, like those used in Phasor, are very susceptible to noise.
In fact, we use a bilateral filter to reduce the effect of shot noise on the output of Phasor.
Although this effectively reduces random noise, any misalignment on raw correlation measurements (like the one occurring in case of motion) creates a systematic noise that cannot be eliminated by bilateral filtering, \new{which}\old{and} dramatically reduces the accuracy of Phasor (Fig.~\ref{fig:errorMPISmallMotion_example}c). Our MRM appears much more reliable in this situation (Fig.~\ref{fig:errorMPISmallMotion_example}e).

The case of a real scene of a corner, acquired in static conditions with a Kinect 2, is depicted in Fig.~\ref{fig:errorMPIReal_example}.
The ground truth shape of the scene could be estimated by checkerboard calibration applied to each of the three planes of the corner.
This figure shows that MRM can significantly reduce MPI artifacts (compared to LF2) not only in simulation, but also in realistic conditions.

\subsection{Motion Artifact Correction and Ablation Study}

\begin{wrapfigure}{r}{0.64\textwidth}
\begin{minipage}[tu]{0.62\textwidth}
\centering
\includegraphics[width=0.49\textwidth, clip=true, trim=0cm 0cm 1.5cm 1cm] {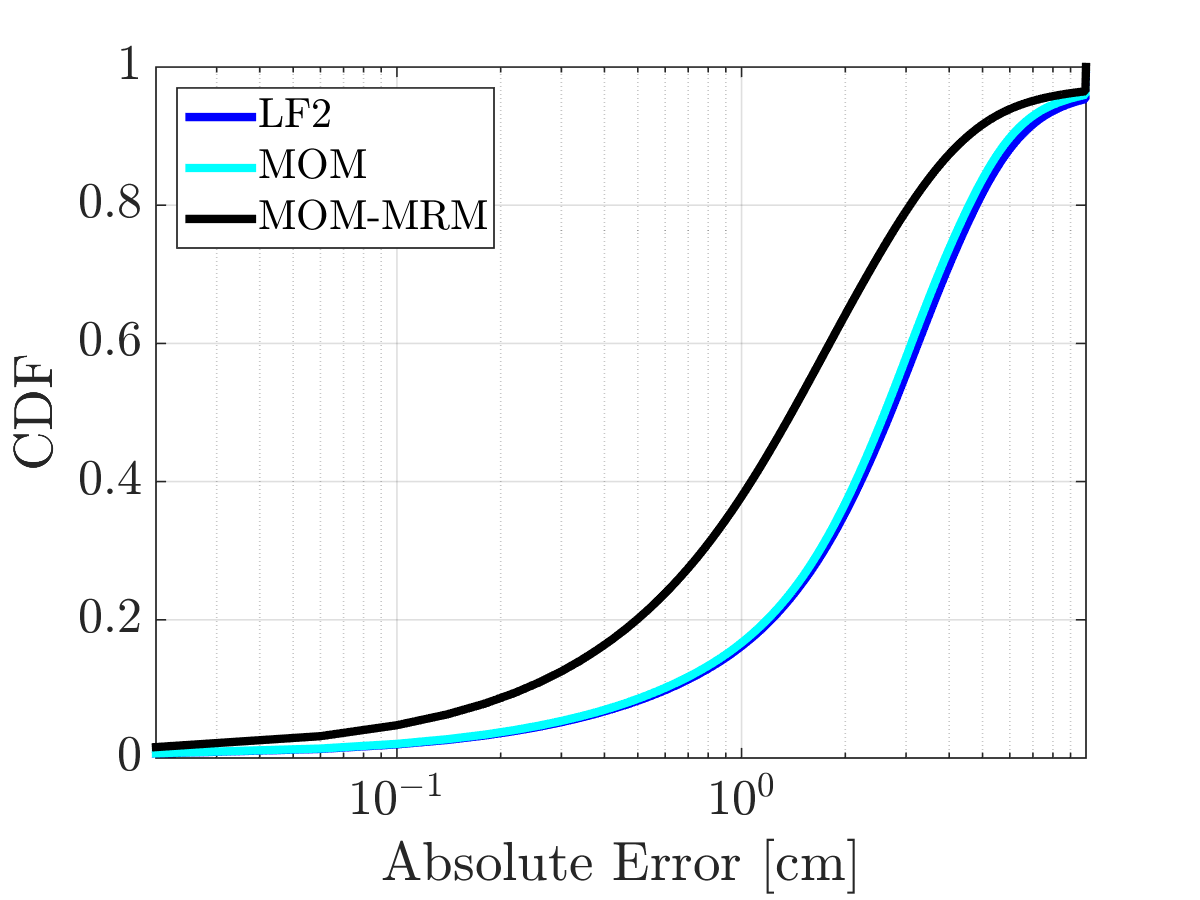}
\includegraphics[width=0.49\textwidth, clip=true, trim=0cm 0cm 1.5cm 1cm]{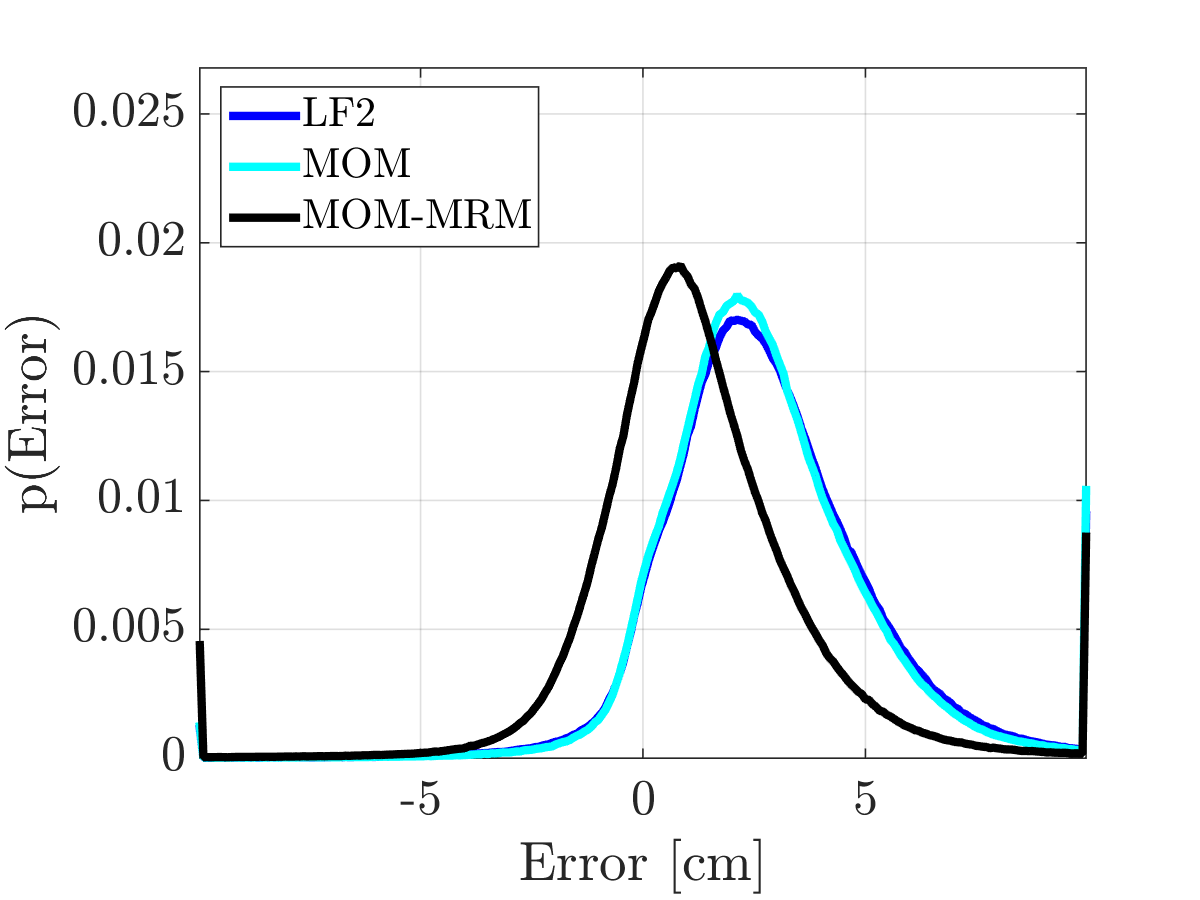}
\end{minipage}
\\
\begin{minipage}[tu]{0.63\textwidth}
\centering
\scriptsize
\begin{tabular}{|c|cc|c|c|}
\hline
  & Median & IQR & Prctile ($90^{th}$) & Density\\
  & \multicolumn{2}{c|}{(error)}& (absolute error) & - \\
\hline
LF2 				& 2.57 (1.11\%) & 2.65 (1.14\%) & 5.62 (2.51\%) & 93.56\%\\
MOM 			& 2.50 (1.08\%) & 2.55 (1.10\%) & 5.48 (2.44\%) & 95.50\%\\
MOM-MRM 	& 1.02 (0.45\%) & 2.43 (1.06\%) & 4.12 (1.82\%) & 97.67\%\\ 
\hline
\end{tabular}
\end{minipage}
\caption{The upper left panel shows the CDF of the depth error for LF2, MOM and MOM-MRM, for simulated data from the FLAT dataset, affected by shot noise, MPI, and motion. The upper right panel shows the histogram of the error. Only those pixels that have been reconstructed contribute to the statistics. The table shows the corresponding reconstruction errors (median, $90^{th}$ percentile, and Inter Quartile Range (IQR), in cm). The numbers in the brackets indicate the relative errors.}
\label{fig:simulated_motion_mpi_shot}
\end{wrapfigure}

\begin{figure}
    \begin{minipage}{\textwidth}
    \begin{center}
    \subfloat[True] {\includegraphics[height=0.127\textwidth]{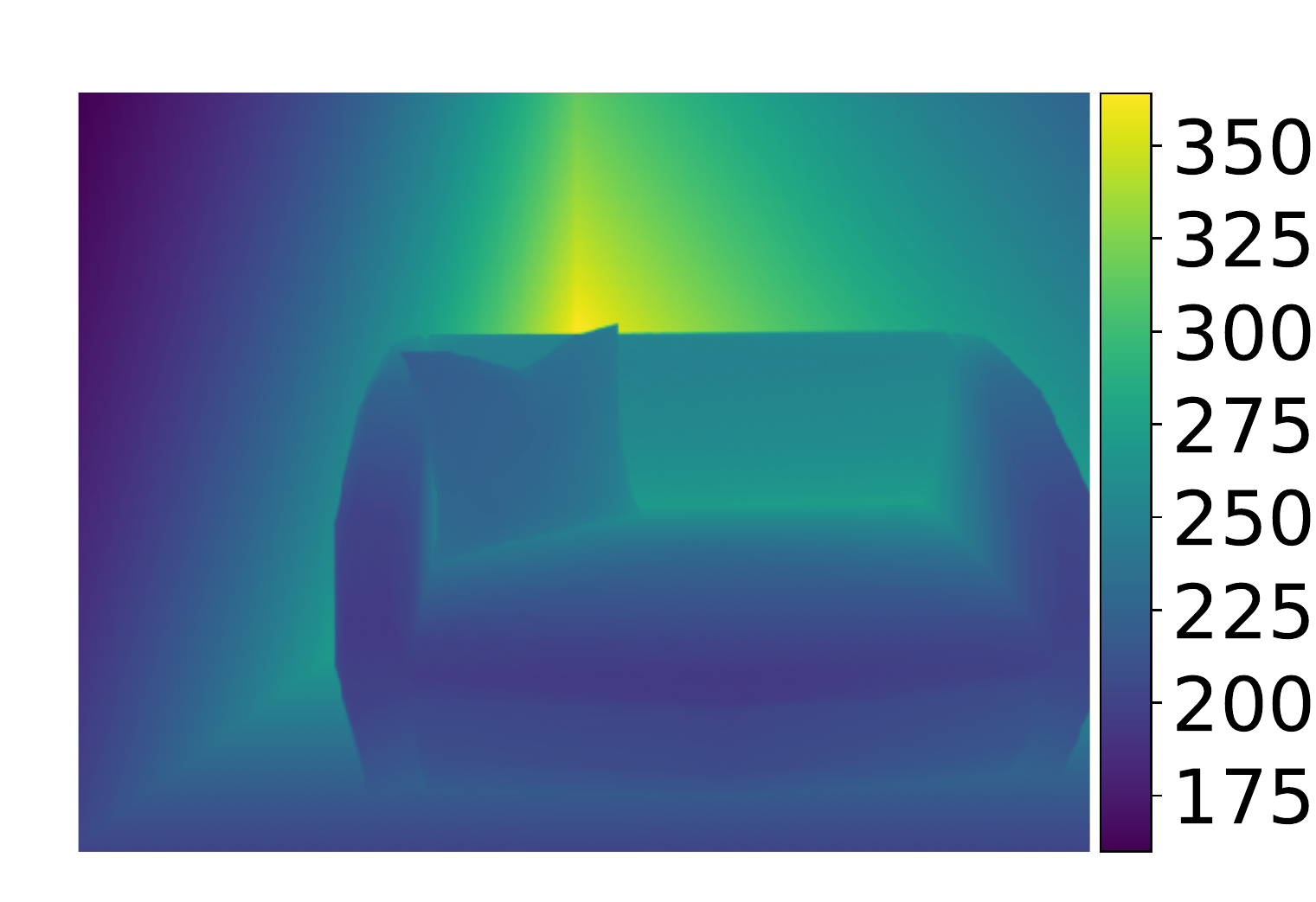}}
    \subfloat[DToF]{\includegraphics[height=0.127\textwidth]{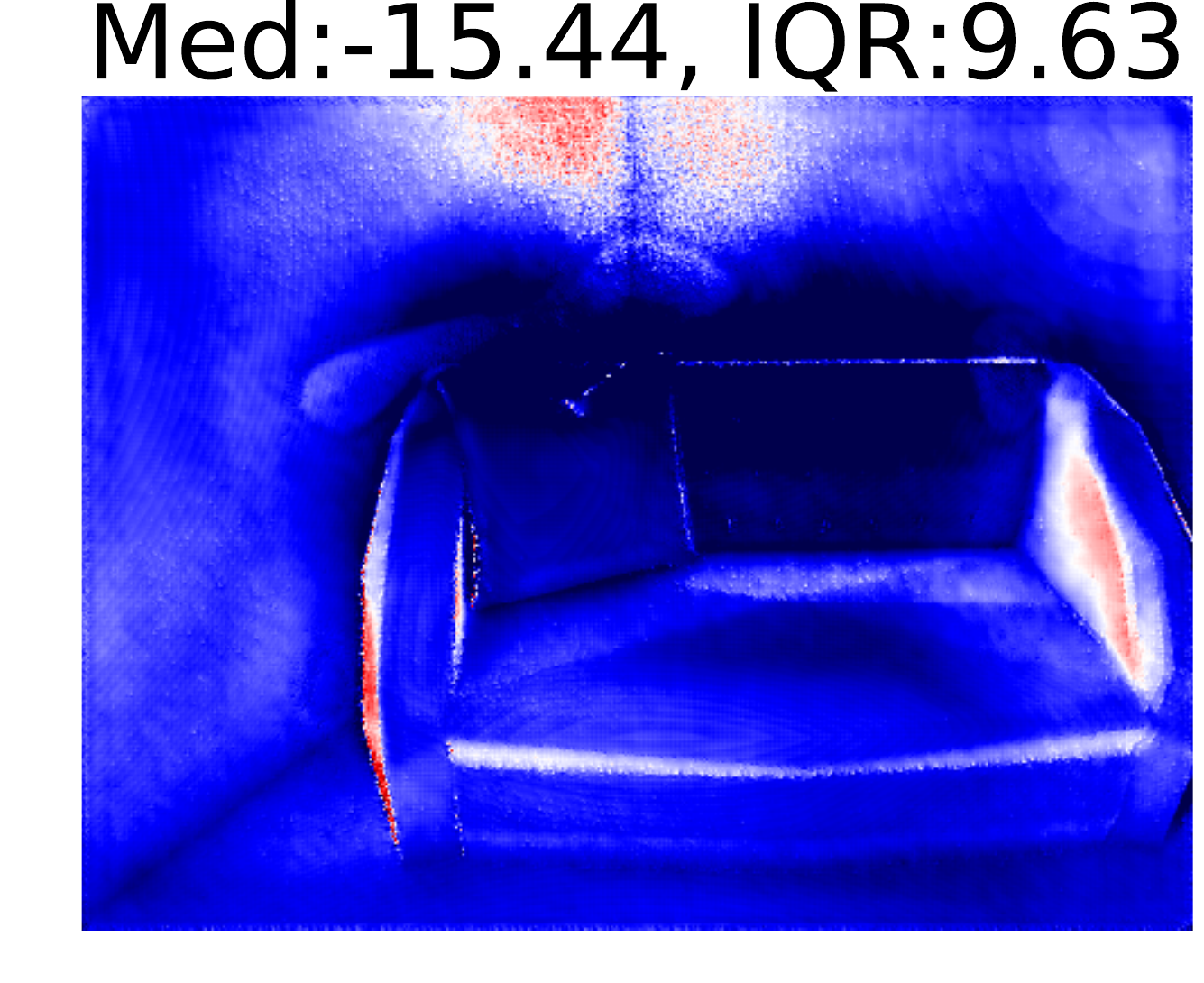}}
    \subfloat[Phasor]{\includegraphics[height=0.127\textwidth]{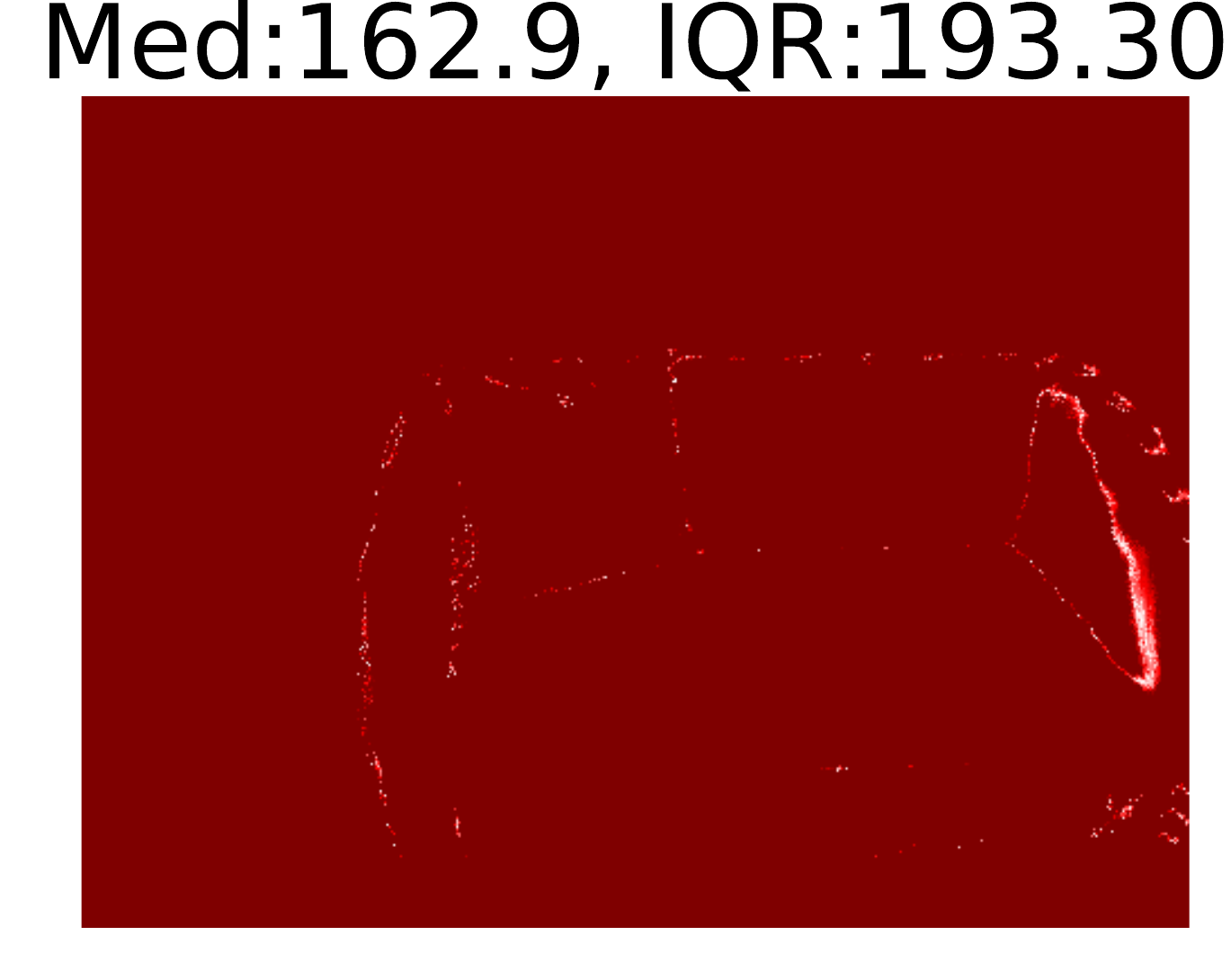}}
    \subfloat[LF2]{\includegraphics[height=0.127\textwidth]{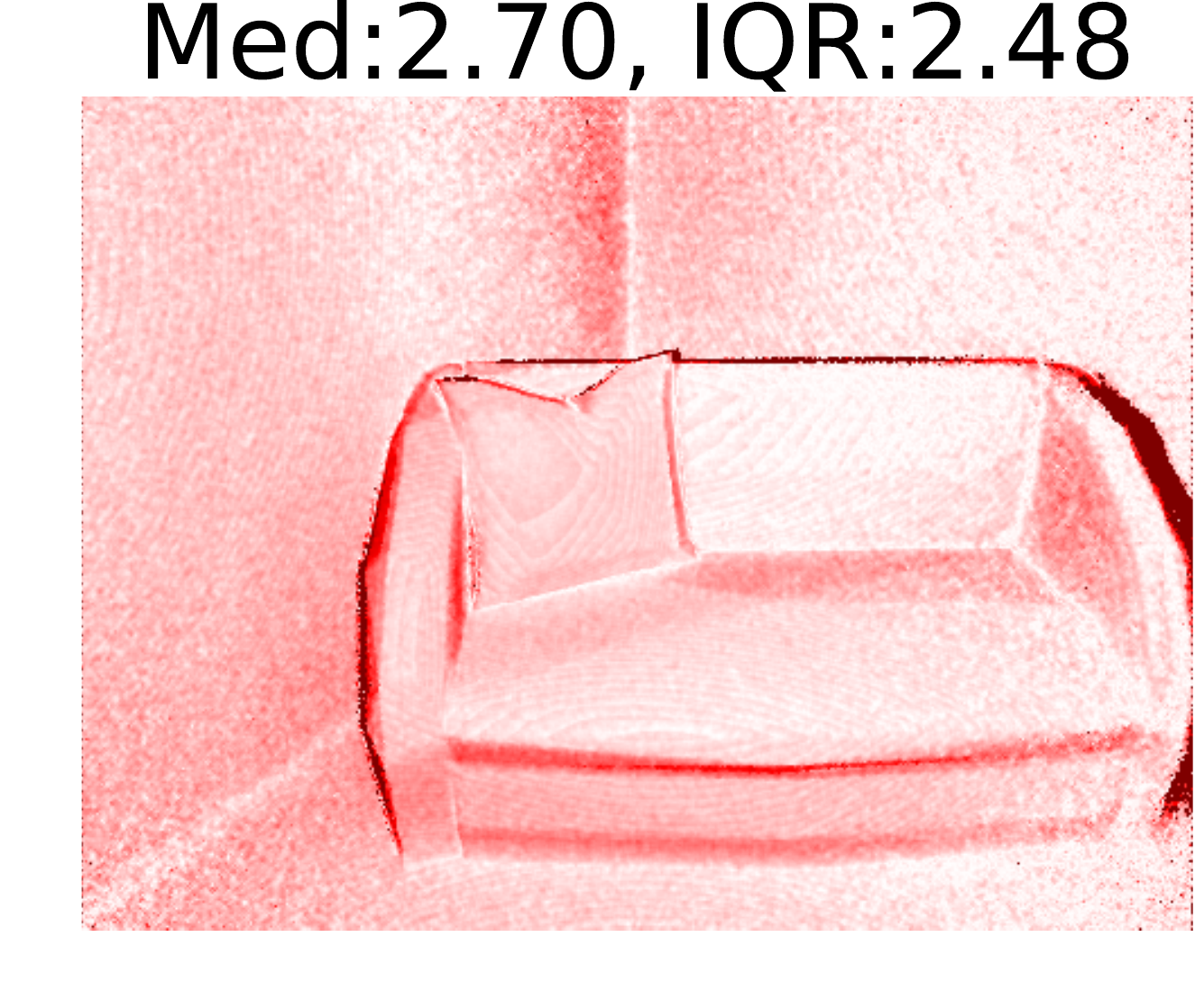}}
    \subfloat[MRM]{\includegraphics[height=0.127\textwidth]{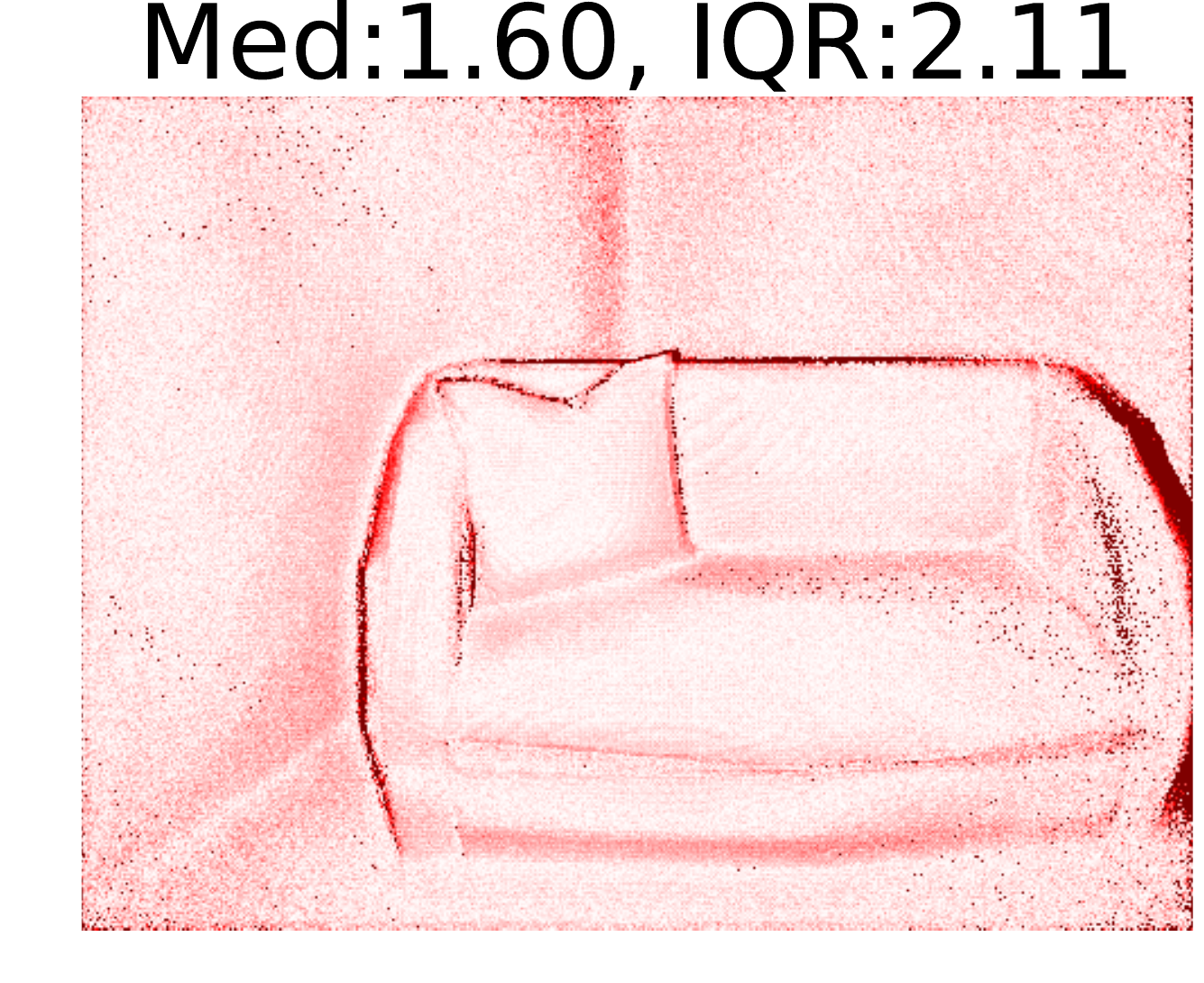}}
    \subfloat[MOM-MRM]{\includegraphics[height=0.127\textwidth]{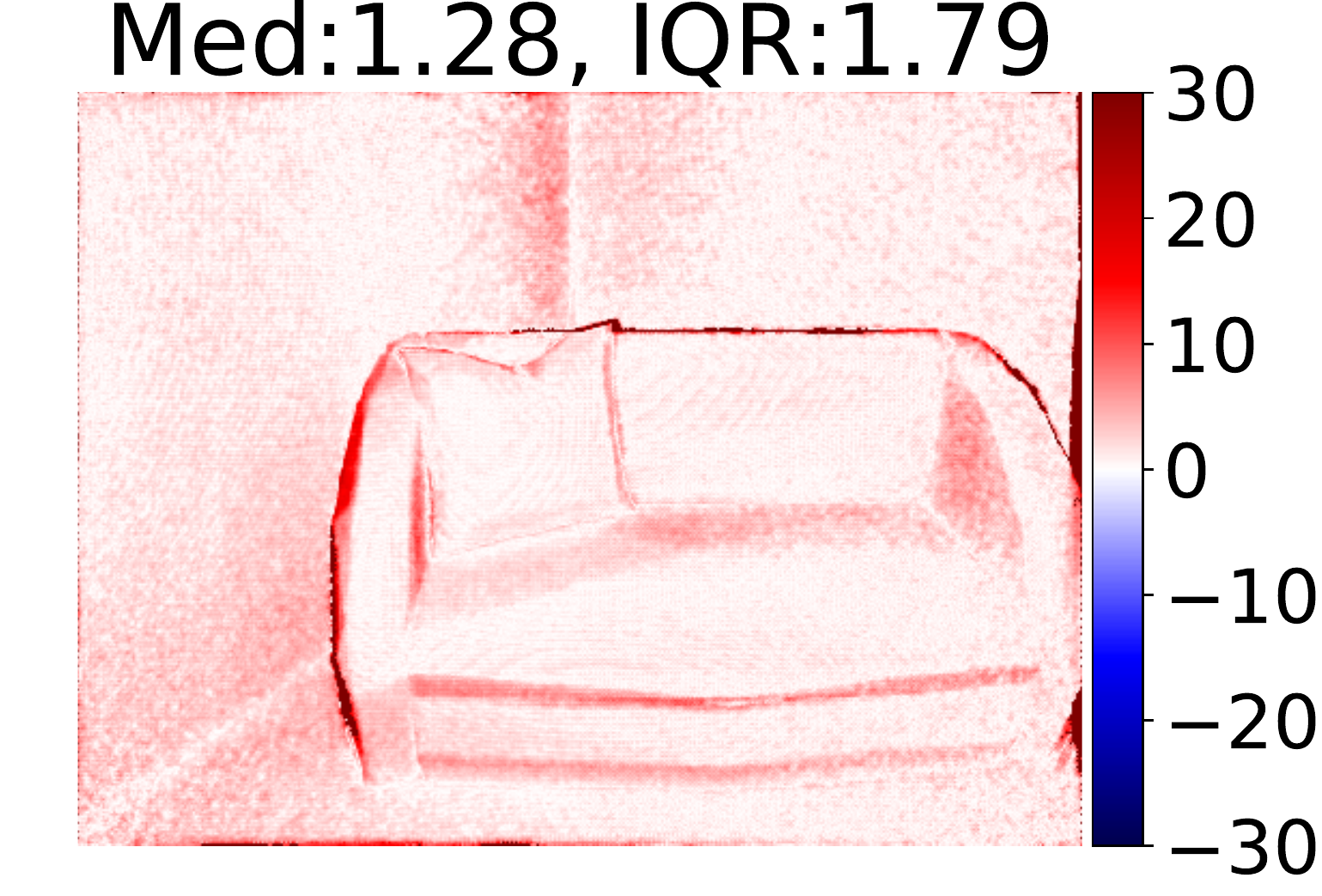}}\\
    \captionof{figure}{Depth error for different reconstruction methods and a simulated scene from FLAT, corrupted by shot noise, MPI, and small motion. Units in cm.}
    \label{fig:errorMPISmallMotion_example}
    \end{center}
    \end{minipage}
    
    \begin{minipage}{\textwidth}
    \centering
    \subfloat[Scene] {\includegraphics[height=0.127\textwidth]{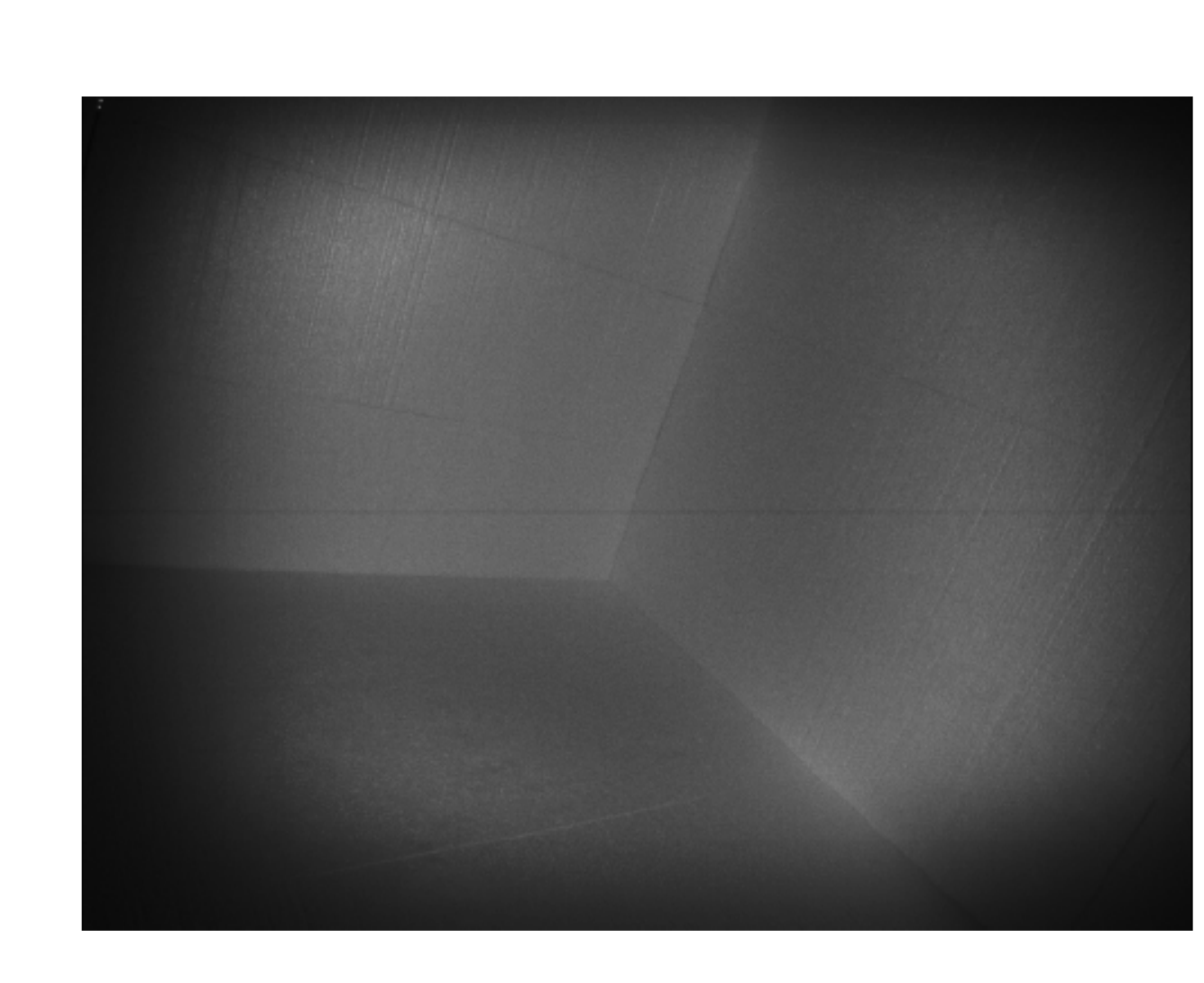}}
    \subfloat[True]{\includegraphics[height=0.127\textwidth]{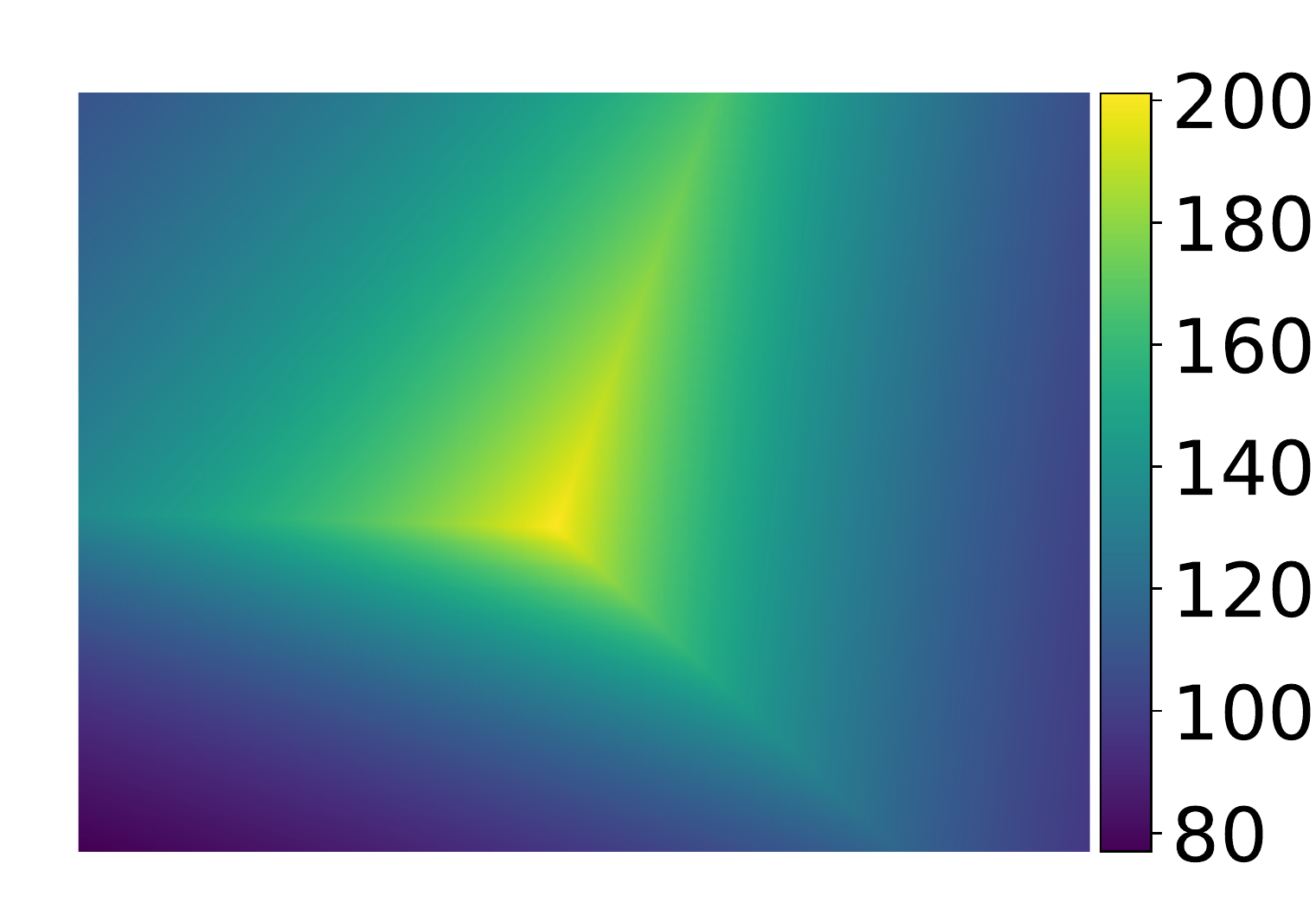}}
    \subfloat[LF2]{\includegraphics[height=0.127\textwidth]{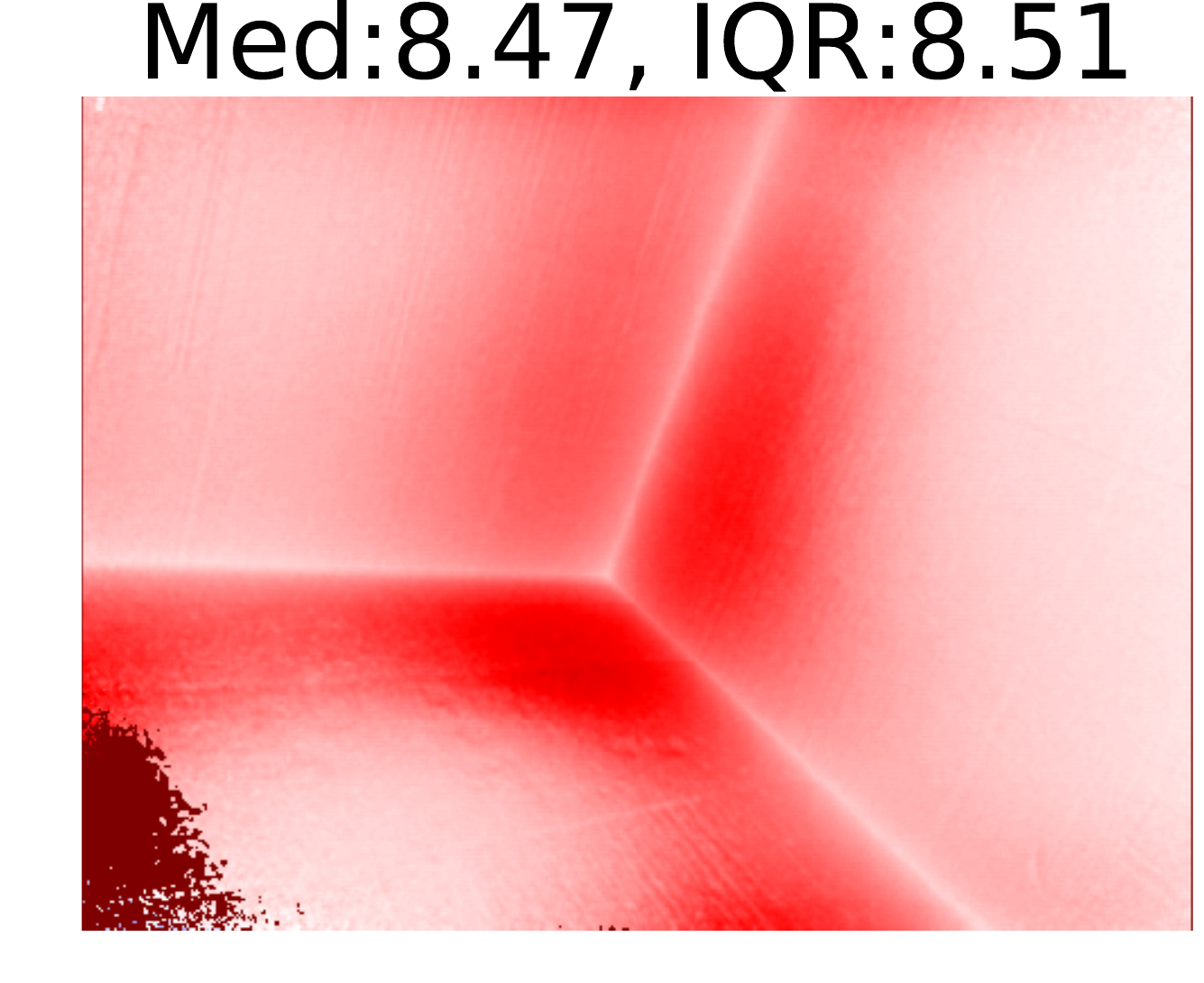}}
    \subfloat[MRM]{\includegraphics[height=0.127\textwidth]{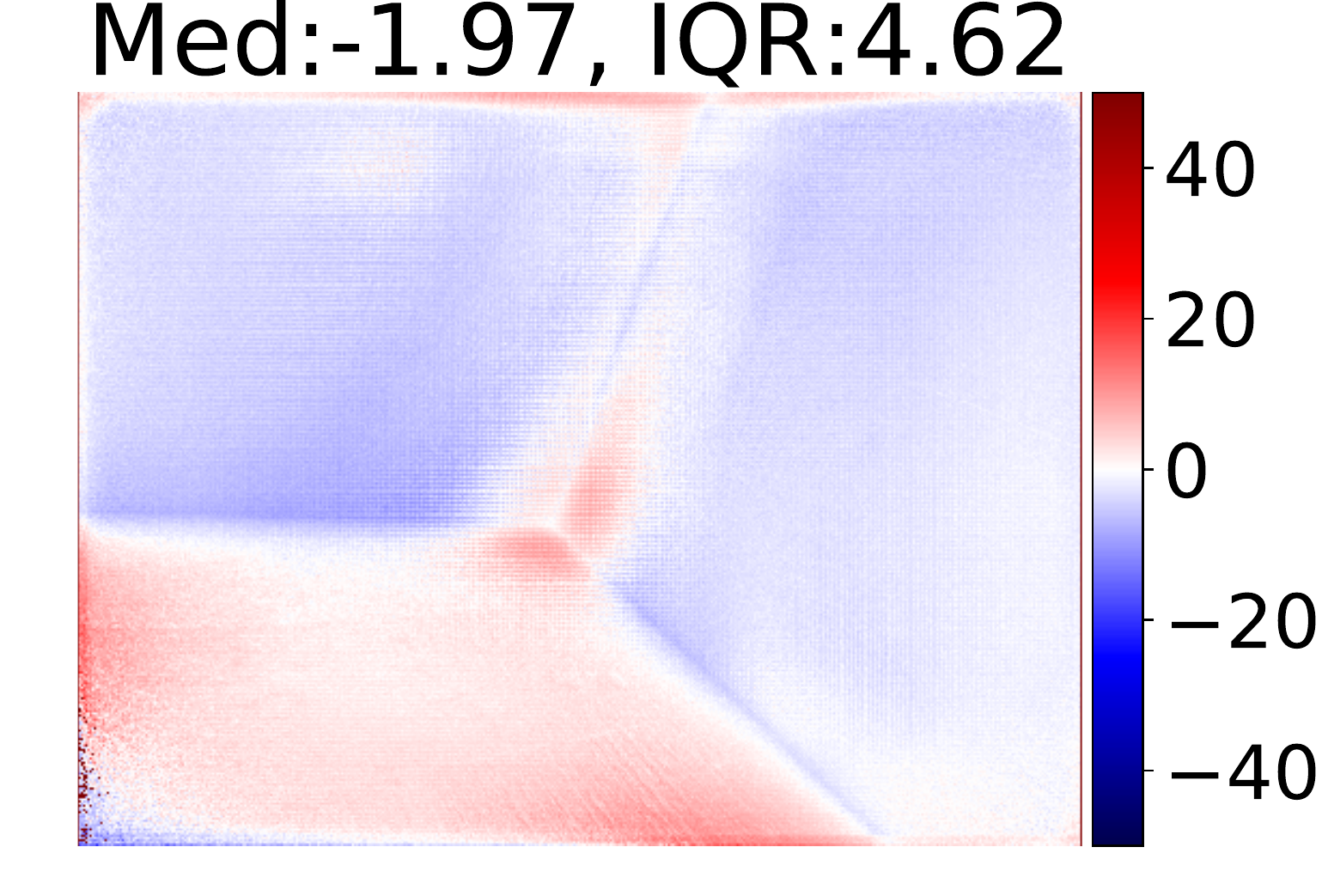}}
    \captionof{figure}{A real scene of a corner captured by a Kinect 2 (a) with ground truth depth (b) and depth errors for LF2 (c), and MRM (d). MPI artifacts  show up as lobes close to the corner in LF2, and are reduced by MRM. Units in cm.}
    \label{fig:errorMPIReal_example}
    \end{minipage}
    
    \begin{minipage}{\textwidth}
    \centering
    \subfloat[True]{
    \begin{tabular}{c}
        \includegraphics[height=0.127\textwidth]{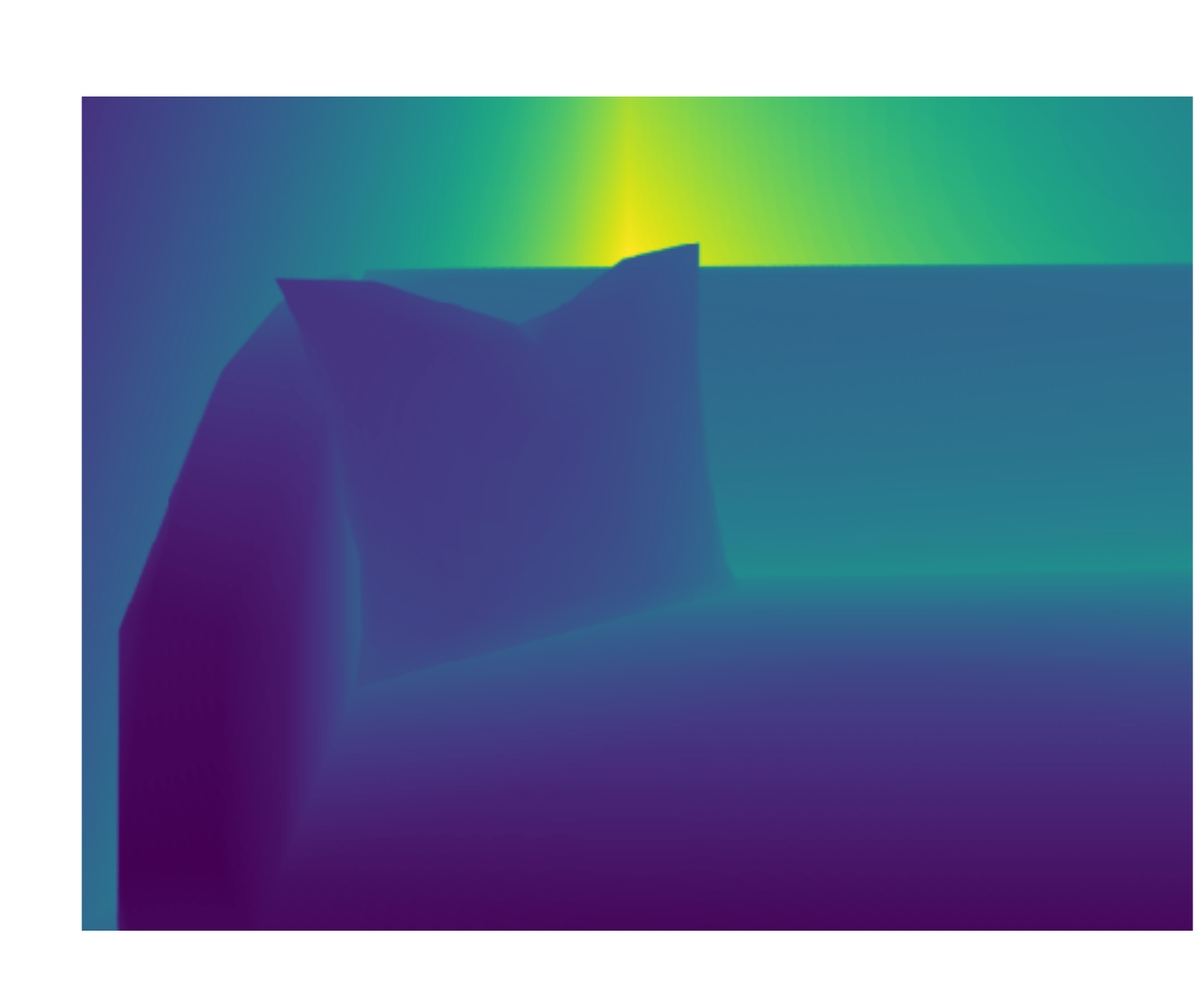}   \\
        \includegraphics[height=0.127\textwidth]{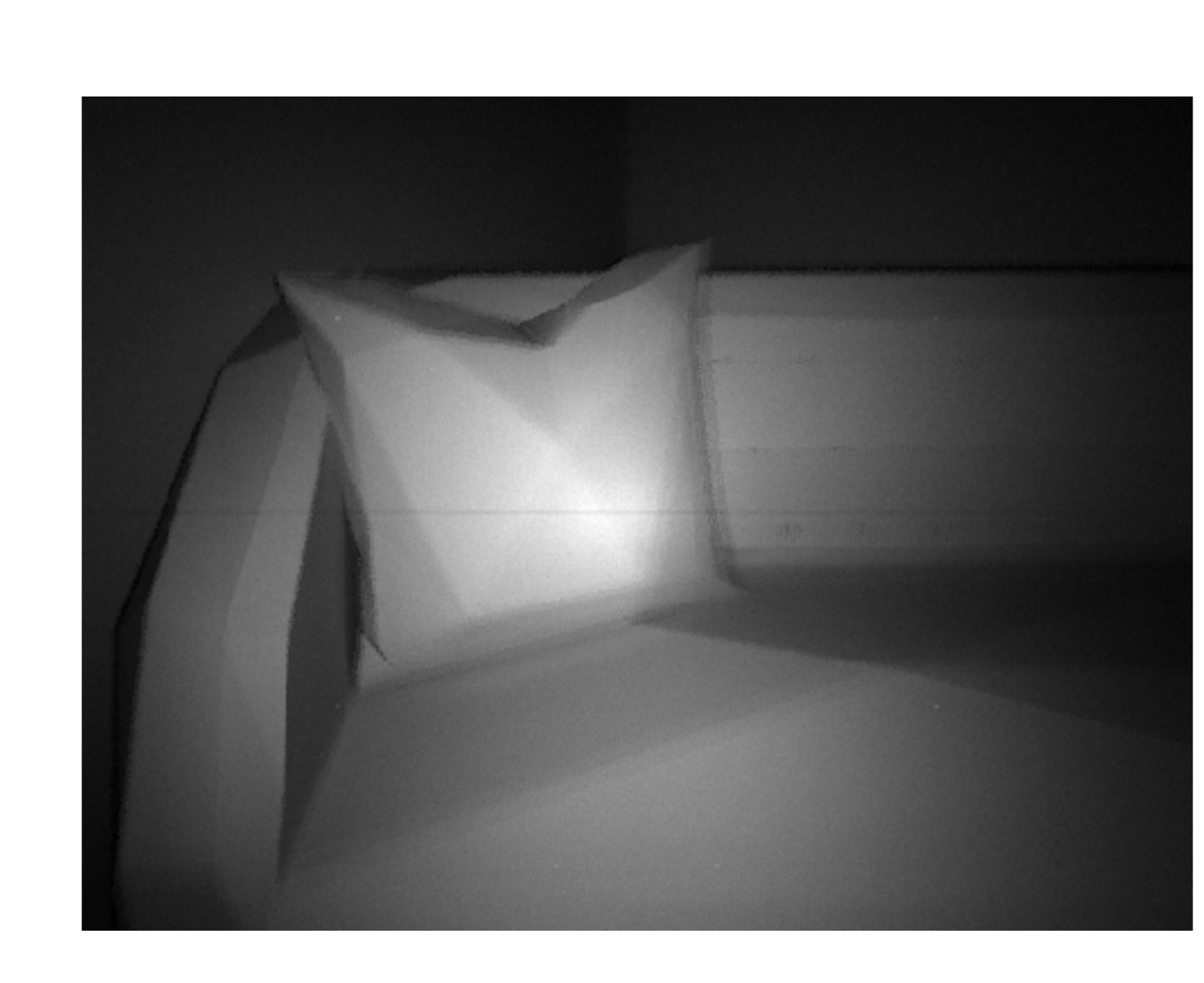}  
    \end{tabular}
    }
    \subfloat[LF2]{
    \begin{tabular}{c}
        \includegraphics[height=0.127\textwidth]{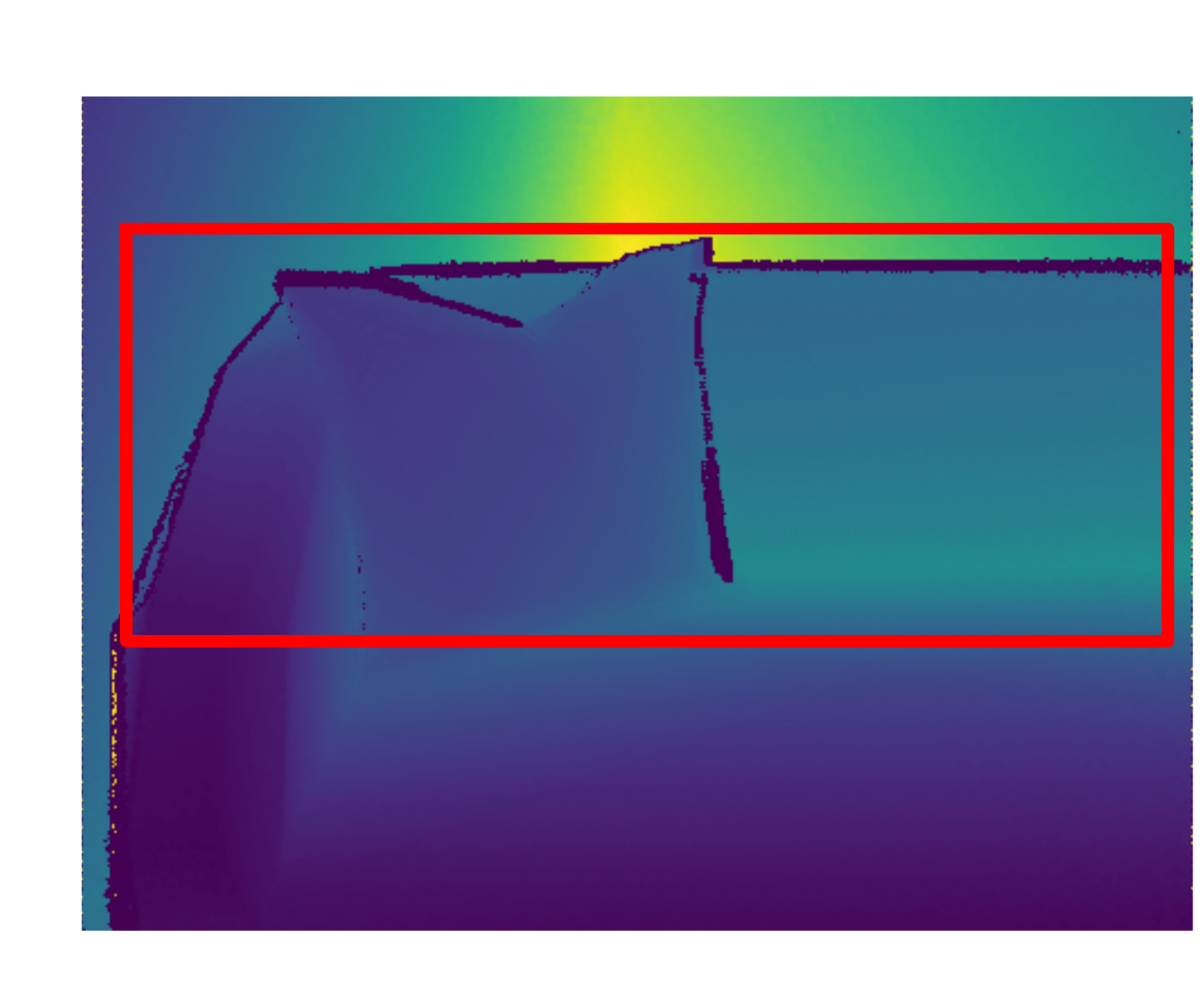}   \\
        \includegraphics[height=0.127\textwidth]{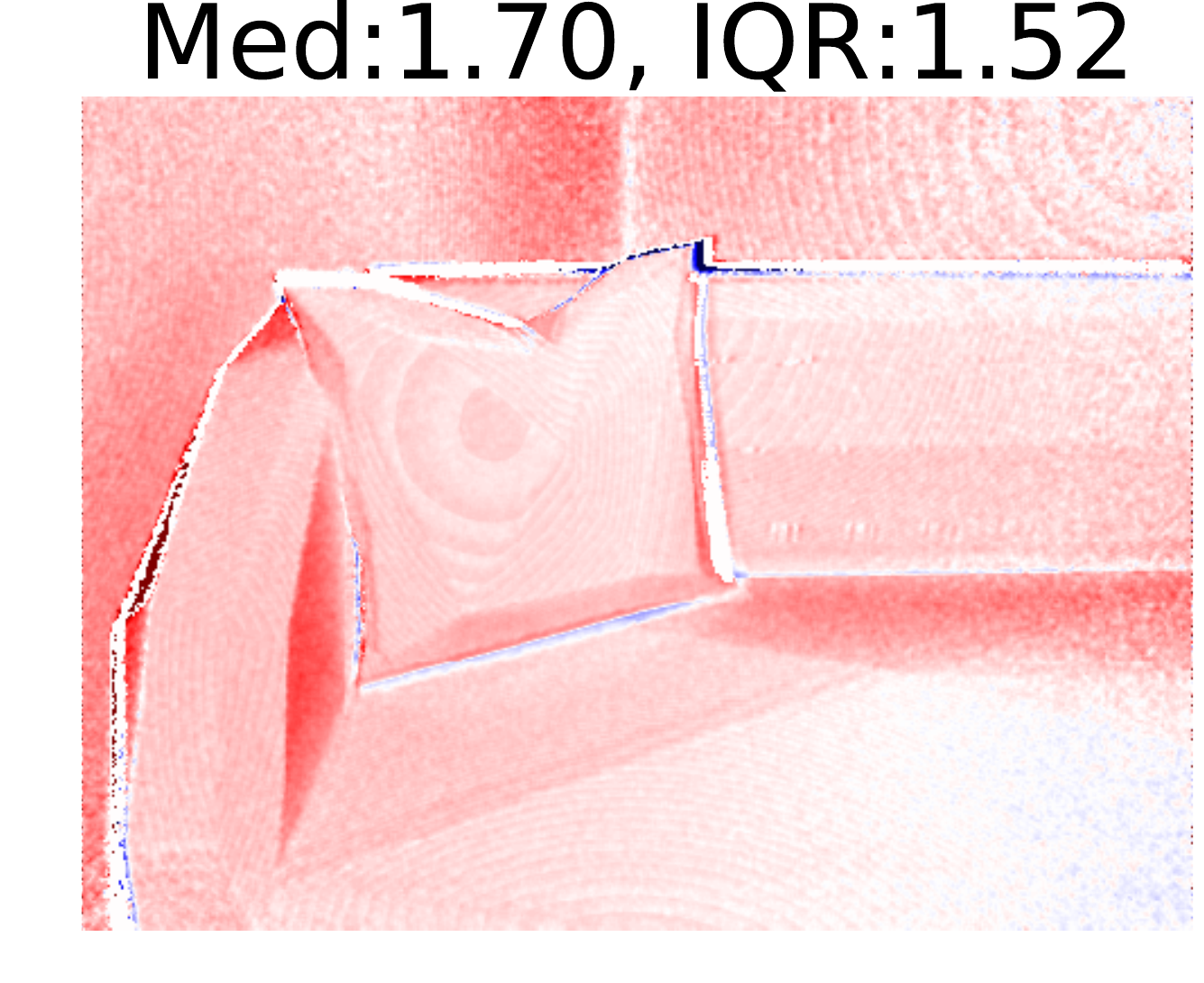}  
    \end{tabular}
    }
    \subfloat[MOM]{
    \begin{tabular}{c}
        \includegraphics[height=0.127\textwidth]{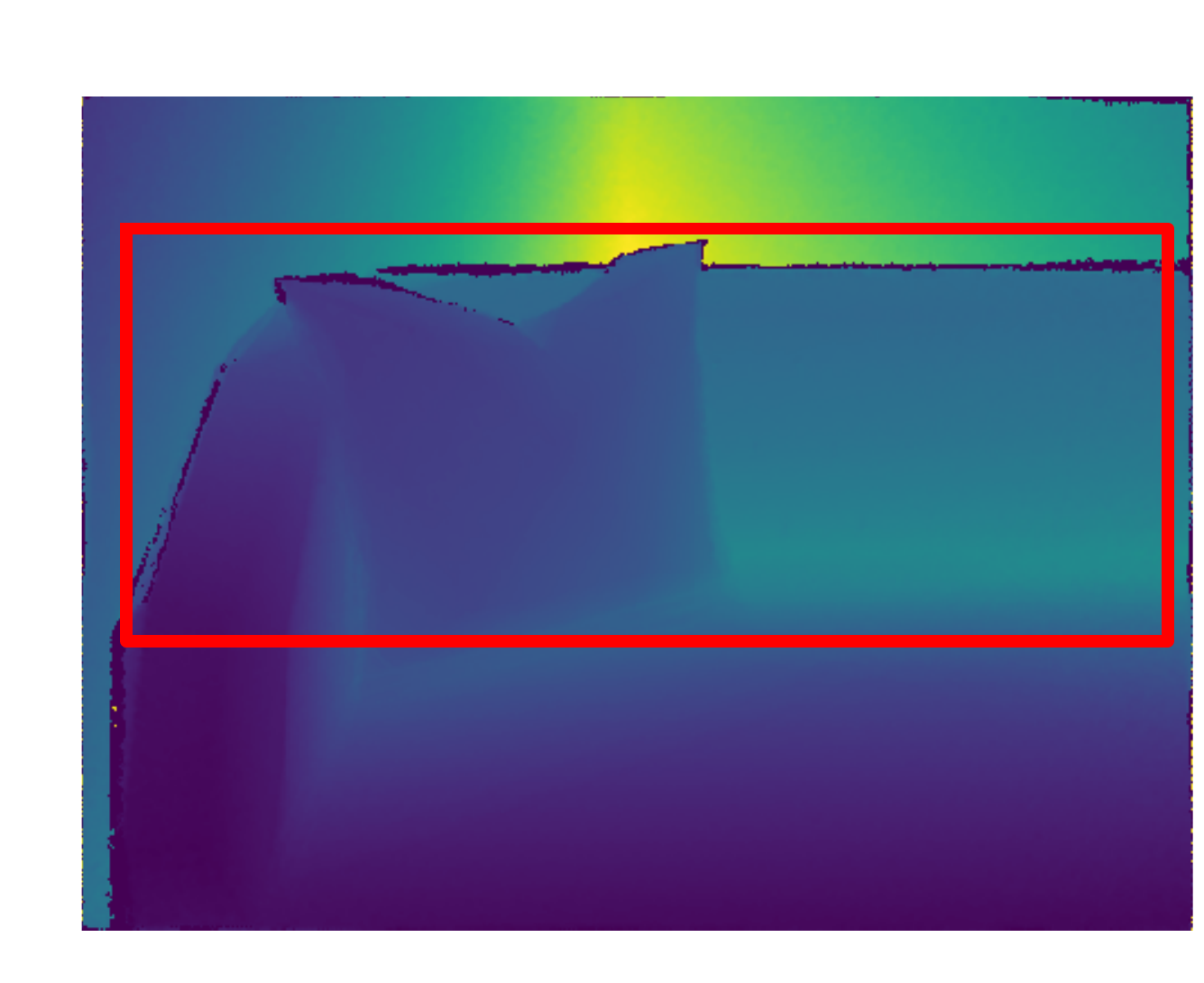}   \\
        \includegraphics[height=0.127\textwidth]{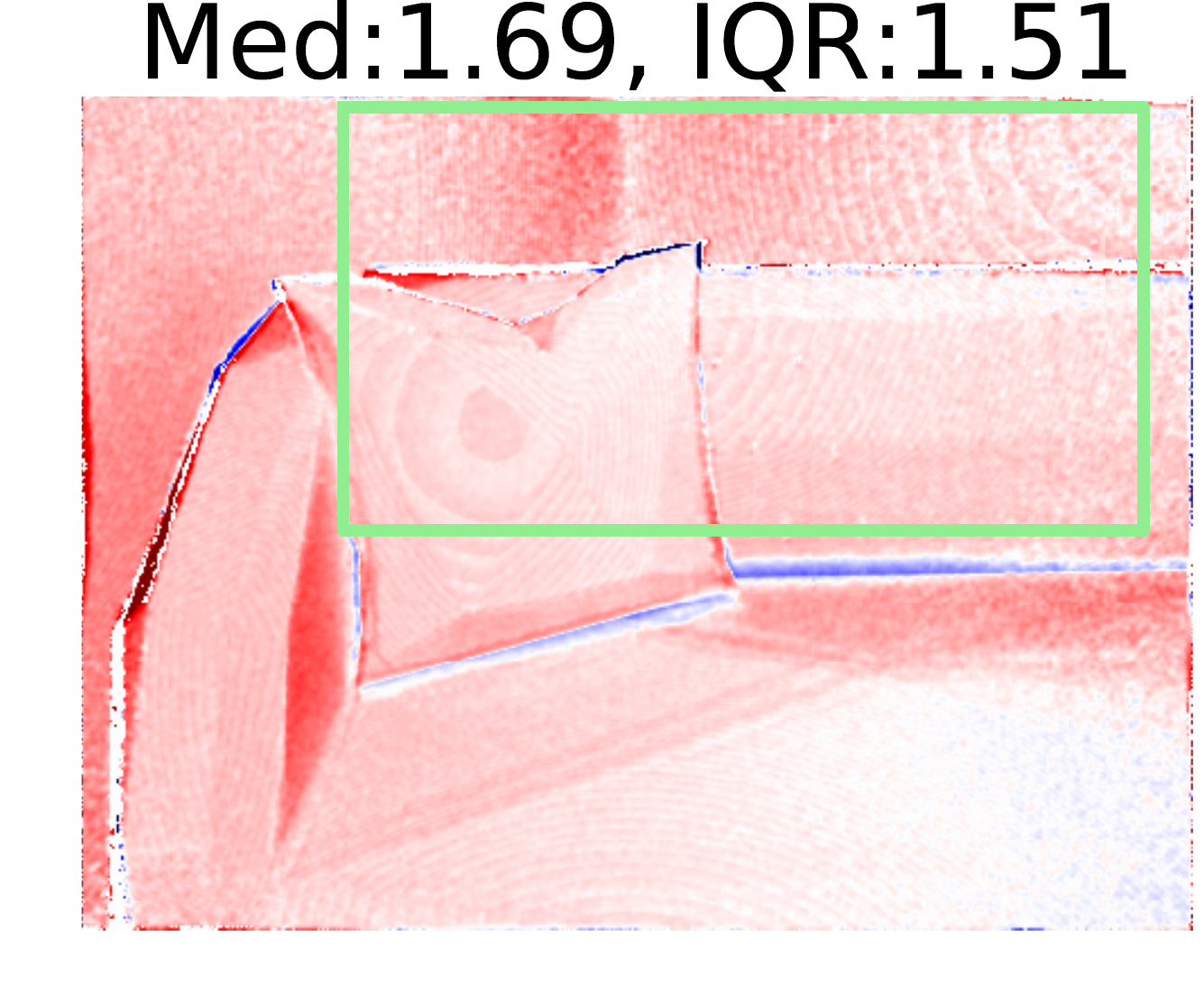}  
    \end{tabular}
    }
    \subfloat[MOM-MRM]{
    \begin{tabular}{c}
        \includegraphics[height=0.127\textwidth]{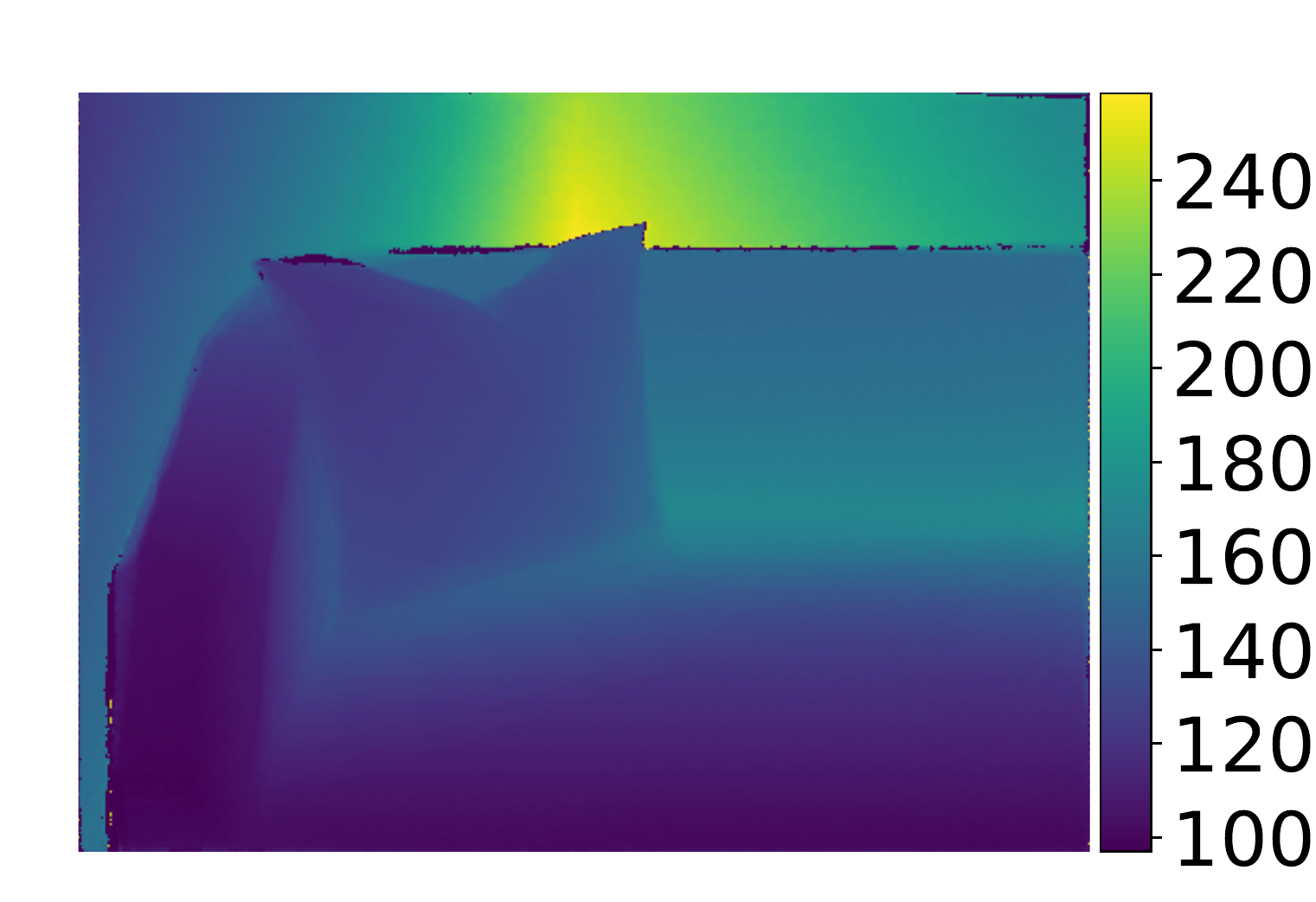}   \\
        \includegraphics[height=0.127\textwidth]{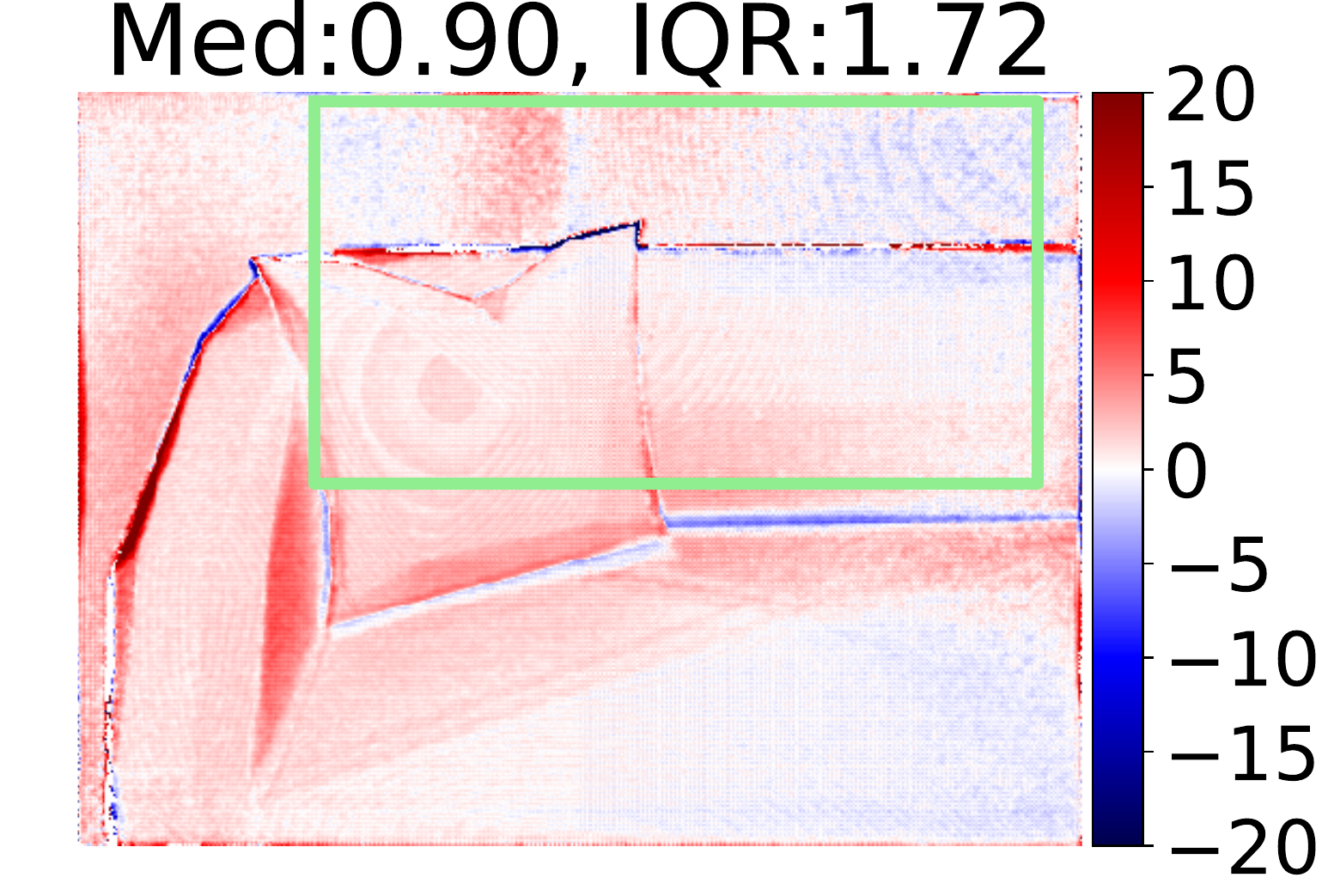}  
    \end{tabular}
    }
    \captionof{figure}{A simulated scene from FLAT, corrupted by shot noise, MPI, and motion.
    The upper row shows the depth maps.
    The lower left panel is the intensity image, other panels in the row are depth errors. 
    MOM aligns object boundaries and allows a more dense reconstruction ({\color{red} red} boxes in b, c).
    MRM mostly corrects MPI artifacts in the smooth areas ({\color{Green} green} boxes in c, d). Units in cm.}
    \label{fig:errorMotion_example_sim}
    \end{minipage}
    
    \begin{minipage}{\textwidth}
    \centering
    \subfloat[Scene]{\includegraphics[height=0.127\textwidth]{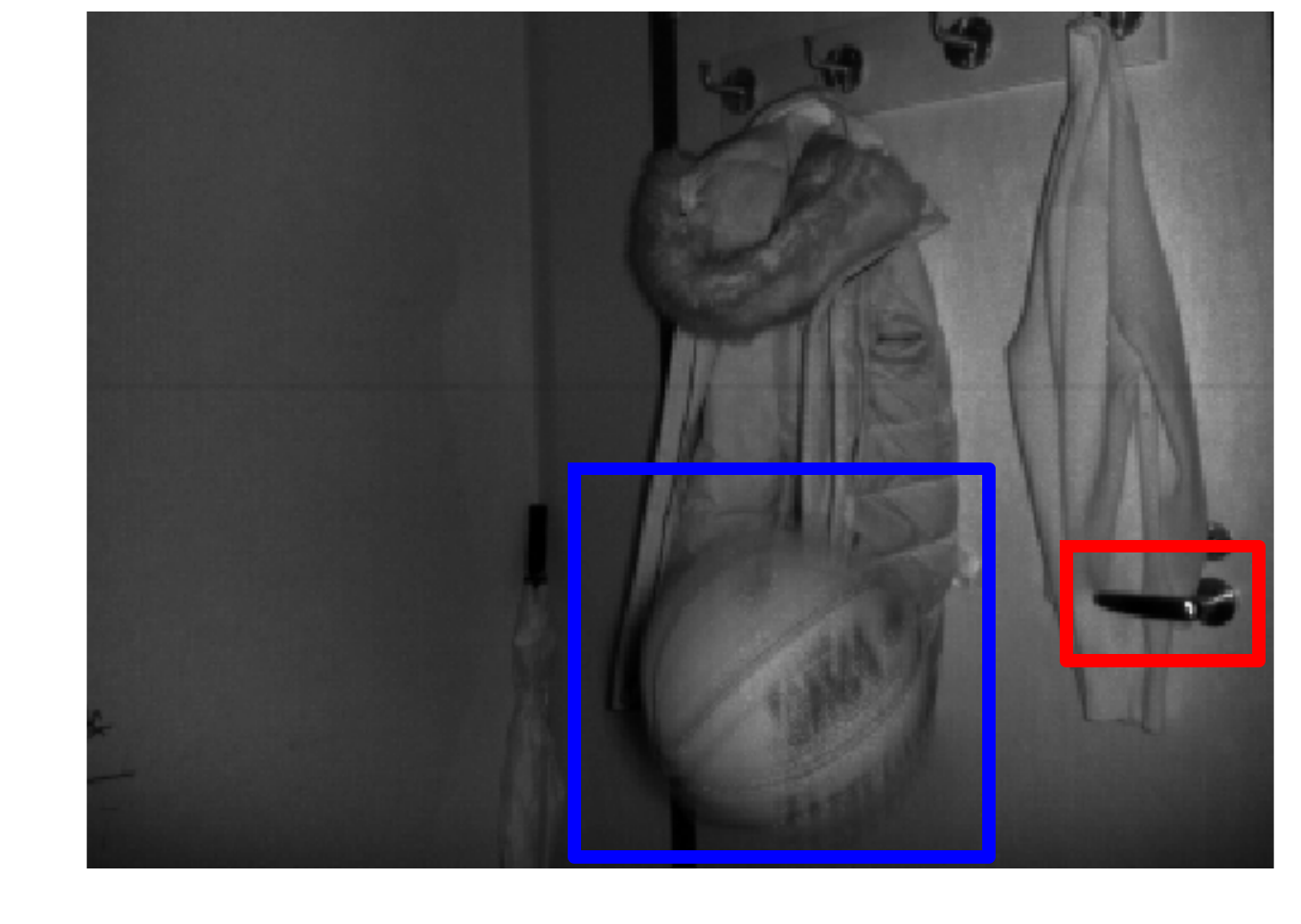}}
    \subfloat[Raw]{\includegraphics[height=0.127\textwidth]{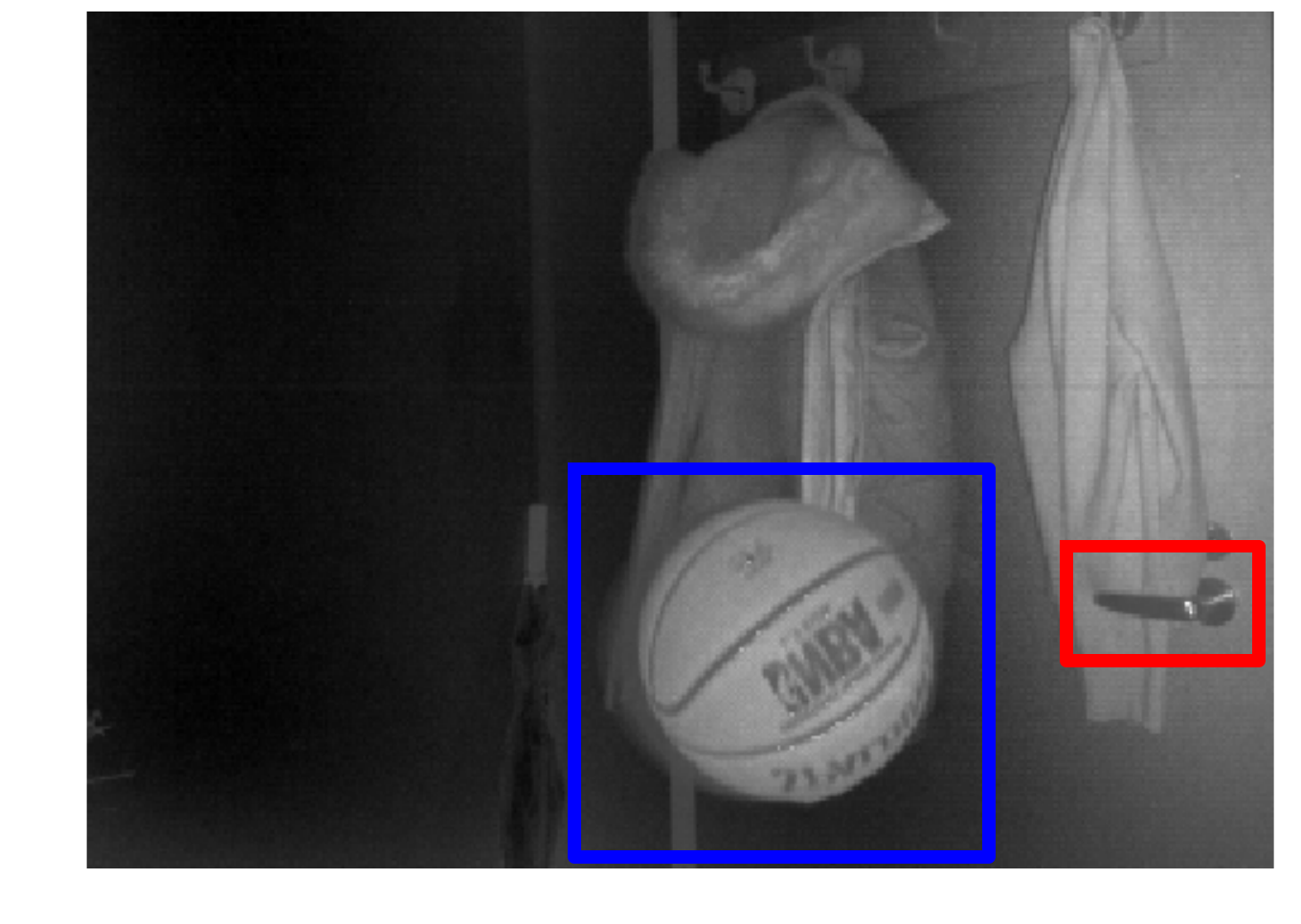}}
    \subfloat[LF2]{\includegraphics[height=0.127\textwidth]{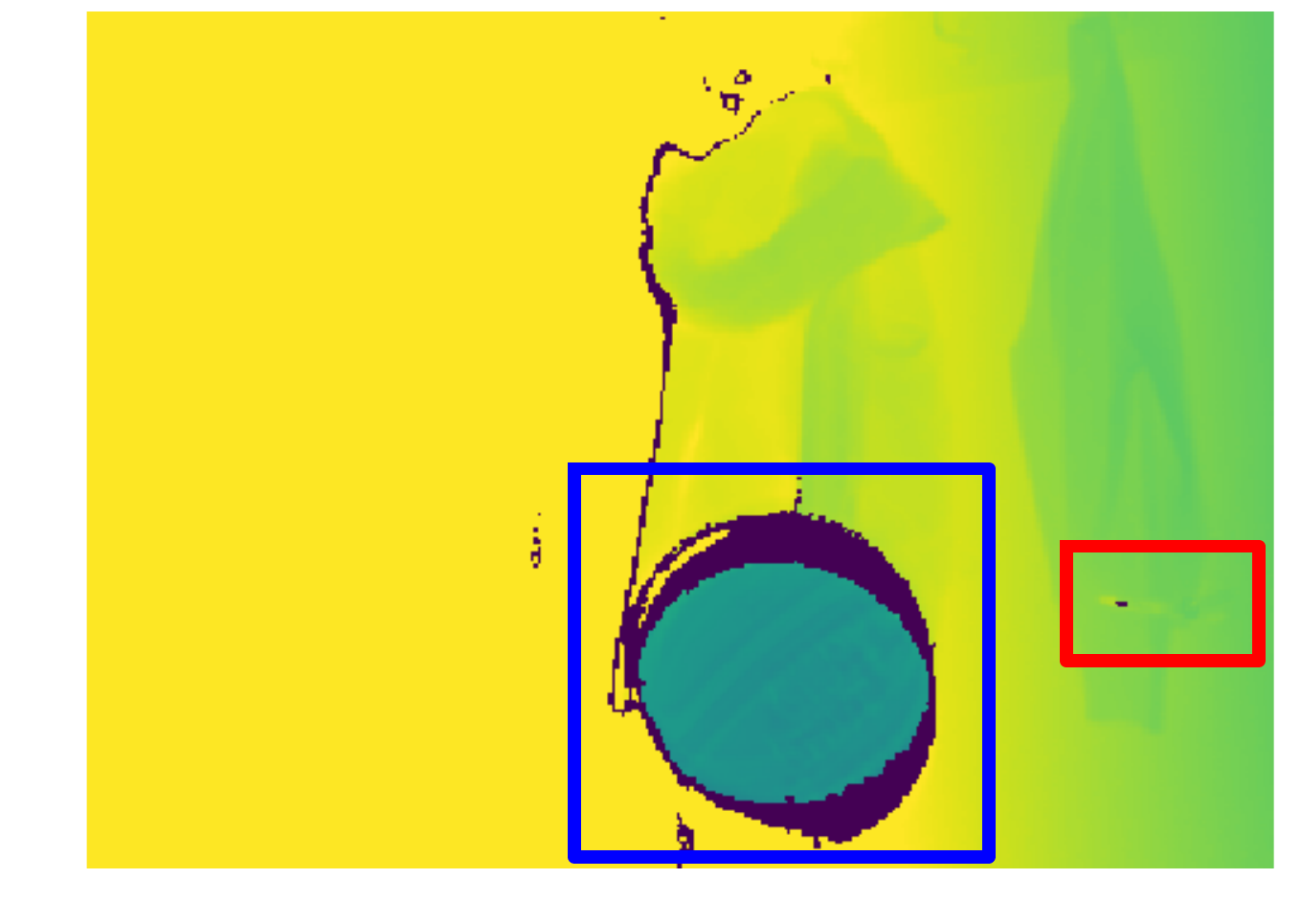}}
    \subfloat[MOM-MRM]{\includegraphics[height=0.127\textwidth]{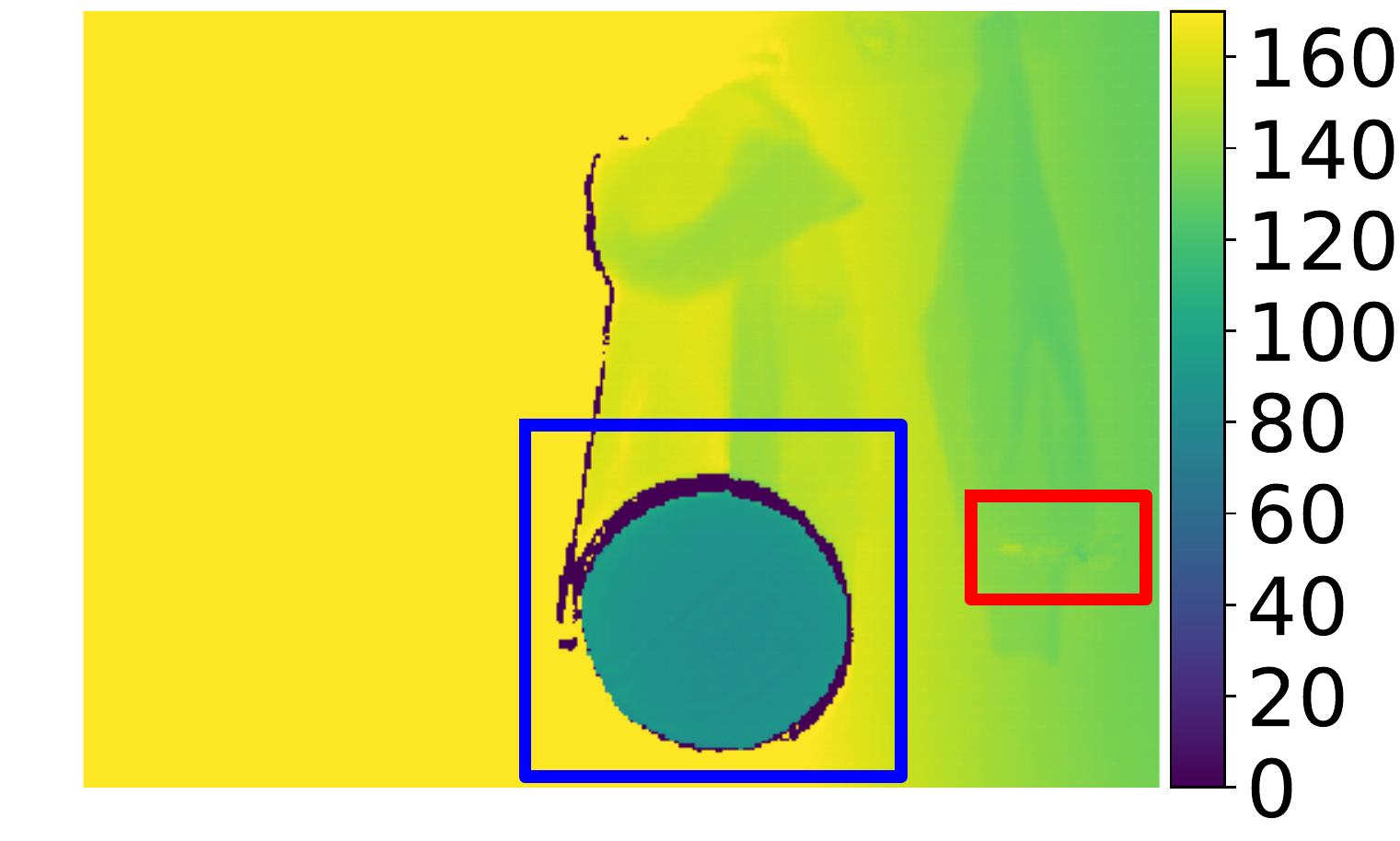}}
    \subfloat[Flow]{\includegraphics[height=0.127\textwidth]{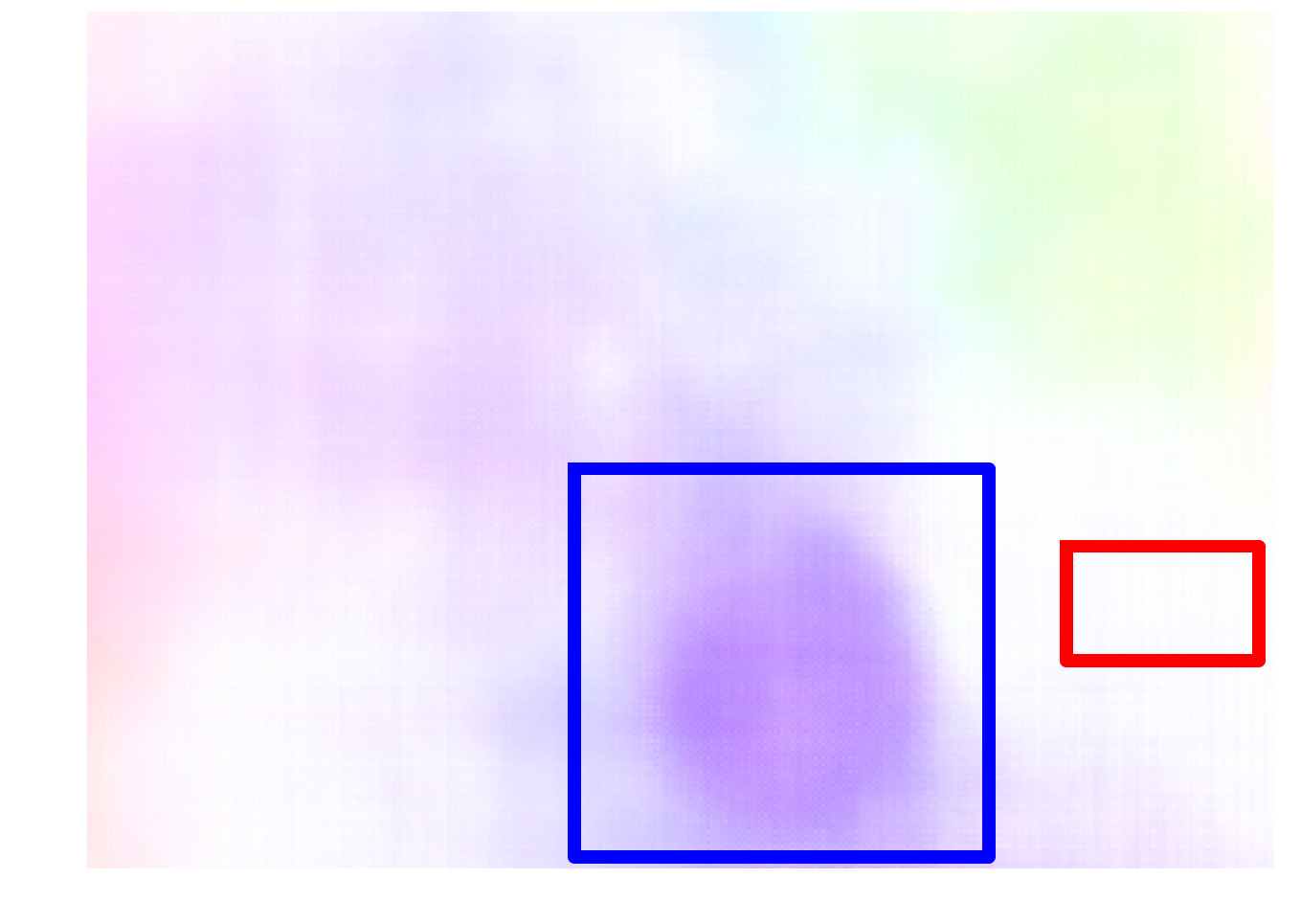}}
    \captionof{figure}{Panel (a) shows the average of nine raw correlation measurements acquired by Kinect 2, with a moving ball ({\color{Blue}blue} box); panel (b) shows one of the raw measurements. Our method (d) reduces the motion artifacts compared to LF2 (c). The optical flow generated from MOM is  shown in panel (e). The {\color{Red} red} box highlights the reflective hand bar with specular reflections; if not masked out, our method fails on these pixels. Units in cm.}
    \label{fig:realMotion}
    \end{minipage}
\end{figure}

We perform an ablation study to quantify the benefits of MOM and MOM-MRM, on a test set from the FLAT dataset corrupted by MPI, shot noise, and random motion.
For this experiment, the motion between the nine correlations measurement is fully simulated by moving the objects in 3D space.
This allows testing the MOM and MOM-MRM on simulated raw correlation measurements affected by a real motion field, even if the modules are trained on approximated motion data. 
We compare depth reconstruction through LF2, MOM, and MOM-MRM, each using the same masking method provided by LF2 to eliminate unreliable pixels, like those along misaligned object boundaries due to motion.
The density (i.e., percentage of reconstructed pixels) reported in Fig.~\ref{fig:simulated_motion_mpi_shot} is 
therefore representative of how well objects boundaries are re-aligned by MOM.
The depth accuracy is slightly higher for MOM compared to LF2, but the main advantage for MOM is a reduction of the unreliable pixels, as density increases from $93.56\%$ to $95.50\%$.
Red boxes in Fig.~\ref{fig:errorMotion_example_sim} demonstrate in simulation how introducing the MOM module can reduce the presence of holes in the reconstructed scene, especially close to object boundaries.
The introduction of the MRM module further increases the density and reduces the bias in the depth error caused by MPI.
Also this effect is clearly visible in \new{the} green boxes in Fig.~ \ref{fig:errorMotion_example_sim}, where the introduction of the MRM module leads to the reduction of the MPI artifact in the corner of the room.

\subsection{Putting Everything Together}
Fig.~\ref{fig:errorMPISmallMotion_example} shows a simulation from the FLAT dataset, where the scene has been corrupted by shot noise, MPI, and a \new{small}\old{tiny} motion \new{(an average of $10$ pixels between the first and last raw correlation measurements).}\old{, with an average motion of about $10$ pixels between the first and last raw correlation measurements.}
Phasor imaging cannot produce a reliable depth map in this case: even a small motion changes the measured phase significantly, because of the fast modulation frequency.
MRM still outperforms LF2, reducing the MPI, whereas the architecture trained to correct both motion and MPI, MOM-MRM, performs\old{at} best (lowest median and IQR). The reconstruction of the depth scene using the MOM-MRM approach takes approximately 90ms on a NVIDIA TitanV GPU.

Our DNN architecture, which operates on raw measurements, is the result of a thorough investigation.
We tested several architectures, including one that directly outputs depth as suggested in the recent work by Su et al.~\cite{Su_2018_CVPR}, but the reconstruction error was consistently larger than that of the proposed DNN.
We believe this is due to the fact a DNN that outputs the depth directly would be forced to learn the non-linear mapping from raw measurements to depth, instead of leveraging the fact that such relation is dictated by physics and known (see the inverse-tangent of Eq.~\ref{eq:gs}).
An additional benefit of working in the raw domain is that, beyond depth, we can estimates uncertainty as in the LF2 pipeline, which is useful, for instance, to mask out bad depth values.

Fig. \ref{fig:realMotion} shows an example of a real scene with moving objects (ball in the blue box), captured by Kinect 2. Notice that Fig.~\ref{fig:realMotion}a is blurred as it averages nine raw measurements, whereas each individual raw measurement is still sharp, as in Fig.~\ref{fig:realMotion}b. Coherently with simulations, the main  effect of MOM is the reduction of the holes close to the boundaries of the moving object. 

\subsection{Method Limitations}
Our method has some limitations that could be overcome with further development.
The receptive field of MRM (blue box in Fig.~\ref{fig:errorMPISimulated_example}) is 72$\times$72 pixels, in theory not large enough to capture global geometric information and correct long-range MPI; furthermore, our loss function naturally emphasizes short-range MPI correction because the signal is substantially stronger for shorter traveling distances.
Nonetheless, MRM does reduce many long-range MPI artifacts (see the windshield of the car in Fig. \ref{fig:errorMPISimulated_example}).
This can be explained with the a priori information about object appearance learned by MRM, or assuming that MRM learns to project measurements corrupted by long-range MPI onto the single-bounce manifold (details in the Supplementary).

Another limitation is that FLAT only includes diffuse materials. Therefore MRM cannot reconstruct surfaces with strong specular reflections, as the door handle (red box) in Fig.~\ref{fig:realMotion}.

Fig.~\ref{fig:realMotion} also highlights the \new{limitations}\old{inadequacy} of MOM for large motions: while MOM effectively warps pixels of the foreground moving object, it is not designed to inpaint missing data of the partially occluded background.
This results in only partial elimination of the motion artifacts for large motion fields.

As our experiments with Kinect 2 and rapid motion showed it to be negligible (Fig.~\ref{fig:realMotion}b), we did not consider blur within a single raw measurement.
In the case motion blur becomes significant for other platforms, approximate blur can be easily included when simulating measurements from FLAT.

Lastly, our method assumes constant ambient light (as in typical indoor conditions) to model the camera noise. Characterizing the noise induced by ambient light separately may lead to a more accurate noise model.

\section{Conclusion}
\label{sec:conclusion}


Motion, MPI and shot noise can significantly affect the accuracy of depth reconstruction in ToF imaging. 
We have shown that deep learning can be used to reduce motion and MPI artifacts jointly, on an off-the-shelf ToF camera, in the domain of raw correlation measurements.
We demonstrated the effectiveness of our MOM and MRM modules through an ablation study, and reported results on both synthetic and real data.
Alternative methods to tackle these artifacts are still to be explored; we believe that our flexible FLAT dataset represents a helpful instrument along this research line. 


\bibliographystyle{splncs}
\bibliography{egbib}
\end{document}